\documentclass[runningheads]{llncs}

\usepackage{eccv}

\usepackage{eccvabbrv}

\usepackage{amsmath,amsfonts,bm}

\def\eqref#1{equation~\ref{#1}}

\def\1{\bm{1}}

\def\vtheta{{\bm{\theta}}}

\def\vn{{\bm{n}}}

\def\vx{{\bm{x}}}

\DeclareMathAlphabet{\mathsfit}{\encodingdefault}{\sfdefault}{m}{sl}
\SetMathAlphabet{\mathsfit}{bold}{\encodingdefault}{\sfdefault}{bx}{n}

\usepackage{graphicx}
\usepackage{amsmath}
\usepackage{amssymb}
\usepackage{booktabs}
\usepackage{fontawesome}
\usepackage{multirow}
\usepackage{pifont}
\usepackage{colortbl}

\usepackage[accsupp]{axessibility}  

\usepackage{hyperref}

\usepackage{cuted}
\usepackage{capt-of}
\usepackage{xspace}
\definecolor{cyan}{cmyk}{.3,0,0,0}
\usepackage{blindtext}

\usepackage{orcidlink}

\makeatletter
\DeclareRobustCommand\onedot{\futurelet\@let@token\@onedot}
\def\@onedot{\ifx\@let@token.\else.\null\fi\xspace}
\def\eg{\emph{e.g}\onedot} 
\def\ie{\emph{i.e}\onedot} 
 
\def\etc{\emph{etc}\onedot}

\makeatother

\let\oldparagraph\paragraph
\renewcommand{\paragraph}[1]{\oldparagraph{\textbf{#1}}}

\usepackage[capitalize]{cleveref}
\crefname{section}{Sec.}{Secs.}
\Crefname{section}{Section}{Sections}
\Crefname{table}{Table}{Tables}
\crefname{table}{Tab.}{Tabs.}

\newcommand{\red}[1]{\textcolor{red}{#1}}

\newcommand{\blue}[1]{\textcolor{blue}{#1}}

\def\ourmethod{{Ours}\xspace}
\newcommand{\cmark}{\ding{51}}%
\newcommand{\xmark}{\ding{55}}%
\newcommand{\scmark}{\textcolor{black}{\ding{51}}{\small\textcolor{black}{\kern-0.7em\ding{55}}}}%

\newcommand{\snowflake}{\protect\includegraphics[width=0.32cm,height=0.32cm]{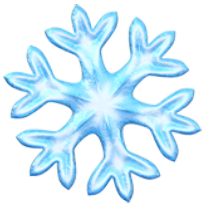}}
\newcommand{\water}{\protect\includegraphics[width=0.32cm,height=0.32cm]{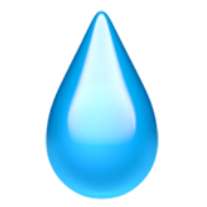}}
\newcommand{\fire}{\protect\includegraphics[width=0.32cm,height=0.32cm]{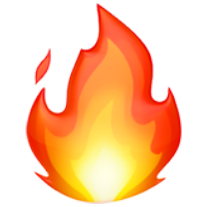}}

\begin{document}

\title{Scene-Conditional \\ 3D Object Stylization and Composition} 


\author{
Jinghao Zhou\orcidlink{0009-0009-6904-3328} \and
Tomas Jakab\orcidlink{0009-0006-5911-9569} \and
Philip Torr\orcidlink{0009-0006-0259-5732}
\and
Christian Rupprecht\orcidlink{0000-0003-3994-8045}
}

\authorrunning{J.~Zhou et al.}

\institute{
University of Oxford \\
\email{\{jinghao,tomj,chrisr\}{\hypersetup{hidelinks}\href{mailto:@robots.ox.ac.uk}{@robots.ox.ac.uk}}}
}

\maketitle

\thispagestyle{empty}
\begin{center}
\centering
\captionsetup{type=figure}
\includegraphics[width=\textwidth]{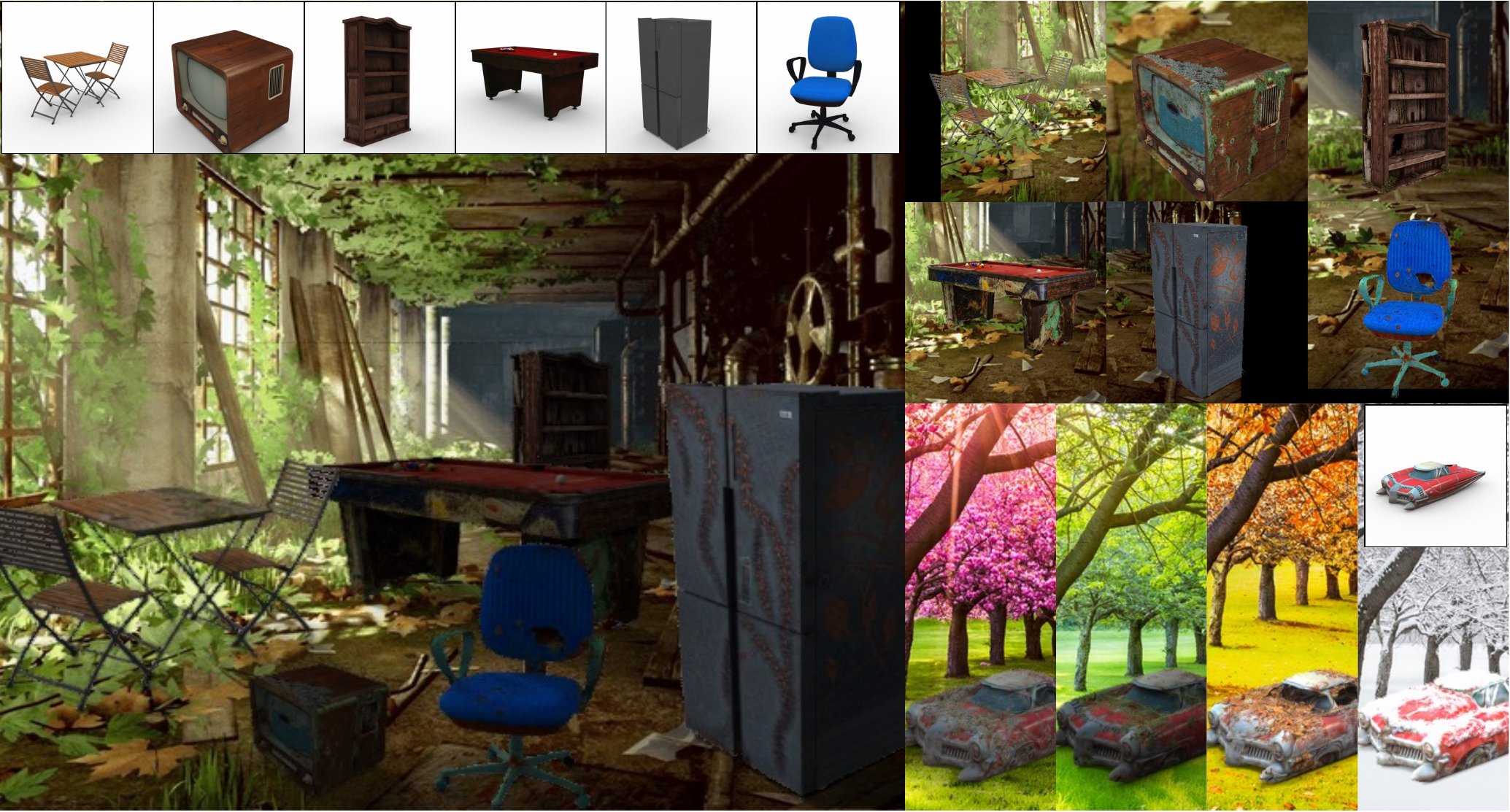}
\captionof{figure}{
We present a framework that adapts a 3D object's appearance to a location in a 2D scene. 
It creates an image where the 3D object is seamlessly blended (left \& bottom right), with its appearance influenced by the scene's environmental conditions and lighting effects.
Moreover, the stylized object with adapted textures (top right), rendered here without the estimated lighting condition for illustrative purposes, can be further utilized as 3D assets for downstream tasks such as video games. 
}
\label{fig:teaser}
\end{center}

\begin{abstract}
\vspace{-0.0cm}
Recently, 3D generative models have made impressive progress, enabling the generation of almost arbitrary 3D assets from text or image inputs. 
However, these approaches generate objects in isolation without any consideration for the scene where they will eventually be placed. 
In this paper, we propose a framework that allows for the stylization of an existing 3D asset to fit into a given 2D scene, and additionally produce a photorealistic composition as if the asset was placed within the environment. 
This not only opens up a new level of control for object stylization, for example, the same assets can be stylized to reflect changes in the environment, such as summer to winter or fantasy versus futuristic settings—but also makes the object-scene composition more controllable. 
We achieve this by combining modeling and optimizing the object's texture and environmental lighting through differentiable ray tracing with image priors from pre-trained text-to-image diffusion models. 
We demonstrate that our method applies to a wide variety of indoor and outdoor scenes and arbitrary objects. See also our
\href{https://shallowtoil.github.io/scene-cond-3d/}{project page}.
\end{abstract}

\section{Introduction}
\label{sec:intro}
After the success of generative models~\cite{dalle,ddpm,stablediffusion} for images in computer vision, the community is now interested in lifting these models to 3D~\cite{dreamfusion,sjc} through advances in architectures, data, and training.
These lifted 3D models demonstrate plausible 3D representations of objects conditioned on images or texts.

While generating whole objects---\eg \textit{an astronaut riding a horse}---is a challenging problem that generative models are becoming able to solve, in practice, these assets are difficult to adapt for use in downstream applications or existing virtual environments such as video games as they are generated without context. 

This paper aims to provide a mechanism to close the appearance gap between 3D objects and environments. We formulate this problem through the lens of a creative tool. 
The intended goal of this paper is to align the appearance of a 3D object to a customized 2D scene as we place it into a specific location, a challenging task that requires fine-grained visual control for convincing results. 

We are interested in producing a photorealistic scene-object composition where the 3D object is seamlessly blended.
For instance, an object placed in a rainy night scene should appear wet and dimly lit, and if the scene is muddy and sunny, the object should also look muddy and cast a strong shadow. To achieve a high level of realism on a fine-grained level, the mud on the object must match the scene's mud in appearance and the direction of the shadow must match where the sun is located. 
Instead of simply resorting to 2D blending operations~\cite{cdc,objectstitch}, we explicitly model an object's appearance and lighting in 3D space, which enables more accurate and regularized control of the object's appearance in the composed image.

We also aim to preserve the underlying object geometry and appearance factors unrelated to the scene environment. 
This task is highly practical; for example, for media design, when presenting a 3D asset of a product within a realistic scene, we want the product to blend naturally while remaining recognizable. 
Manually achieving this would require a skilled artist to modify the object's texture to match the scene, a challenging and time-consuming process.
Our paper introduces a method that automates this task. 
It accepts a 3D object with texture, a 2D scene, and the object's position within the scene and outputs an adapted texture consistent with the scene. 

To accomplish this, we leverage recent advances in generative diffusion models~\cite{stablediffusion,imagen} for images, which are trained on a large-scale dataset of internet images. 
We conduct an optimization procedure where we 
compose the 2D renderings of the 3D object with the 2D scene
at a specified location and use a pre-trained diffusion model to ``critique'' the realism of the composed image, providing a gradient signal for optimization.

The challenge is that diffusion models, such as Stable Diffusion \cite{stablediffusion}, are trained on 2D images and lack any notion of 3D geometry.
Naively optimizing or denoising in 2D space~\cite{objectstitch,cdc} would result in losing the object's 3D geometry.
To prevent this, we optimize the object's content in 3D by randomly rotating the object during optimization and rendering it from various viewpoints, which helps to align the adapted texture with the object's geometry and avoid overfitting to a single viewpoint.

Notably, randomly rotating the 3D object stops lighting effects from being baked into the texture, which is desirable. Yet, it also prevents the object from naturally blending with the scene in a single viewpoint.
To counteract this, we predict the scene's lighting and the shadows cast by the object as separate components.
For this purpose, we propose a novel technique inspired by a common physical approach from the computer graphics industry, where reflective and diffuse spheres with known material parameters are placed and photographed in a real-world scene to capture the environmental lighting conditions.
We adopt this concept and place a virtual white diffuse sphere inside the scene during optimization while conditioning the generative model to expect a white sphere within the scene, thereby capturing the model's interpretation of the scene's lighting in an easily extracted form.

Optimizing the object's appearance in 3D offers further significant advantages: 
The user can change the object's pose without re-running the optimization, as the texture is meaningful from all viewpoints, and view-dependent effects are modeled separately by the lighting component. 
Additionally, the 3D object with the adapted texture can be used as a standalone asset in other downstream tasks (\eg in a video game).


In summary, we present a framework that (1) enables the stylization of a 3D object, which can also be used as a standalone asset, given a 2D scene and its location within it; (2) achieves photorealistic scene-object compositing with the help of a novel mechanism to estimate scene lighting from a single image.

We further verify the effectiveness of our method in extensive experiments, demonstrating that our approach can realistically adapt objects to a diverse set of environments, laying the groundwork for practical applications.

\section{Related Work}
\label{sec:relatedwork}

\begin{table}[!tbp]
    \centering
    \caption{\textbf{Comparison with related work.} \textit{Rep.} denotes representation being 2D (\faSquareO) or 3D (\faCube). 
    \textit{Env. Inf.} denotes \textit{Environmental Influence} and \textit{I.D. Pre.} denotes \textit{Identity Preservation}.
    Parameters for the lighting and the appearance model are either scene-agnostic constant (\snowflake), scene-conditioned predictable (\water), or learnable (\fire).}
    \vspace{-0.0cm}
    \label{tab:relatedwork}
    \resizebox{0.98\textwidth}{!}{%
    \setlength{\tabcolsep}{0.8mm}{
    \begin{tabular}{lcccccccc}
    \toprule
    \multirow{2}{*}{Method} & \multicolumn{3}{c}{Scene} & \multicolumn{4}{c}{Object} & \multirow{2}{*}{Compose} \\
    \cmidrule(lr){2-4}\cmidrule(lr){5-8}
    & \rotatebox{0}{\textit{Rep.}} & {Lighting} & \rotatebox{0}{Shadow} & \rotatebox{0}{\textit{Rep.}} & {Appearance}  & \rotatebox{0}{\textit{Env. Inf.}} & \rotatebox{0}{\textit{I.D. Pre.}} &  \\
    \toprule
    OST~\cite{objectstitch} & \faSquareO & - & \scmark & \faSquareO & -  & \scmark & \scmark & Blend \\
    CDC~\cite{cdc} & \faSquareO & Ambient \snowflake & \scmark & \faCube & Diffuse \snowflake & \scmark & \scmark & Blend \\
    \midrule
    IGAN~\cite{immersegan} & \faSquareO & LDR \water & \cmark & \faCube & Diffuse \snowflake & \xmark & \xmark & Copy \& Paste \\
    EVL~\cite{everlight} & \faSquareO & HDR \water & \cmark & \faCube & Diffuse \snowflake & \xmark & \xmark & Copy \& Paste \\
    \midrule
    PrDr~\cite{prolificdreamer} & - & Ambient \snowflake & \xmark & \faCube & Diffuse \fire & \cmark & \xmark & - \\
    Fan3D~\cite{fantasia3d} & - & HDR \snowflake & \xmark & \faCube & PBR \fire & \cmark & \xmark & - \\
    Inst3D~\cite{instruct3dto3d}  & - & Point \snowflake & \xmark & \faCube & Diffuse \fire & \scmark & \cmark & - \\
    \midrule
    \bf \ourmethod & \faSquareO & HDR \fire & \cmark & \faCube & PBR \fire & \cmark & \cmark & Copy \& Paste \\
    \bottomrule
    \end{tabular}}}
    \vspace{-0.5cm}
\end{table}

The purpose of this paper is to adapt the appearance of a 3D object to align with a customized 2D scene.
This solves a more complicated problem than prior works, which often only tackle individual components, including but not limited to 2D scene-object compositing, light estimation, and texturing, but never as a whole. We compare with these methods next and detail their differences in~\cref{tab:relatedwork}.

\paragraph{Scene-Object Compositing.} 
Scene-object compositing is a challenging task that requires a delicate balance between visual transfer and control. 
It has been extensively researched~\cite{bargainnet,dovenet,ssh,composeadv}. 
Recent methods condition the denoising process of diffusion models~\cite{ddpm, song2020score} on object images. 
OST~\cite{objectstitch} aims to seamlessly integrate a 2D object image into a 2D scene by modifying the keys and values in the attention blocks.
CDC~\cite{cdc} considers composing the 2D renderings of a 3D object with a 2D scene by directly modifying denoised latent.

However, these methods rely entirely on the 2D denoising process and do not explicitly model the object's geometry, which often results in a significant loss of structural details and visual identity. This severely hampers these methods' ability to adapt to the environmental and lighting conditions of the scene.

\paragraph{Light Estimation from Single-View Image.}
Estimating light conditions from a single-view 2D image is of practical use in the real world and has been extensively studied for both indoor~\cite{indoor,indoor1} and outdoor scenes~\cite{skymodel,outdoor,outdoor1}.
The distinction between these two settings has been significantly narrowed recently~\cite{immersegan,everlight}, thanks to the outpainting capabilities of diffusion models, which have contributed to a unified framework.
With additional training on large high dynamic range (HDR) datasets such as SUN360~\cite{sun360}, these methods can approximate the location and intensity of the light source, enabling realistic shadows.

However, these methods estimate only the lighting of the scene. They are typically indifferent to how objects should be stylized to match the scene, neglecting the potential environmental influences on the textures of objects.

\paragraph{Mesh Texturing.}

Classic automated approaches~\cite{wei2009state} for mesh texturing are often limited and apply only simple texture patterns.
In contrast, methods~\cite{oechsle2019texture, siddiqui2022texturify, gao2022get3d} based on Generative Adversarial Networks~\cite{goodfellow2014generative} (GANs) are more capable of synthesizing detailed textures. 
Recently, text-conditioned texture generation~\cite{text2mesh, mohammad2022clip} has been made possible through the optimization of CLIP-based objectives~\cite{radford2021learning}.
The latest works now employ powerful text-to-image diffusion models~\cite{ddpm, song2020score}, either through optimization~\cite{magic3d, metzer2023latent, tsalicoglou2023textmesh, prolificdreamer, fantasia3d} using Score Distillation Sampling~\cite{dreamfusion} (SDS), or by wrapping the generated 2D textures onto the mesh~\cite{texture,text2tex,texfusion}.
Importantly, Variational Score Distillation~\cite{prolificdreamer} (VSD) as an improved optimization objective and the adoption of PBR model~\cite{fantasia3d} have greatly boosted the quality of generated textures.
Additionally, an increasing level of requirement for more visual control from users has been partially met~\cite{instruct3dto3d,instructnerf2nerf} through image-conditional diffusion models~\cite{instructpix2pix} to preserve objects' identities.

However, these works focus on texturing the object in isolation from the scene and do not consider the homogeneity of the composition when these assets are placed in certain environments, which limits their practical usage.



\begin{figure*}[!t]
    \centering
    \includegraphics[width=0.9\linewidth]{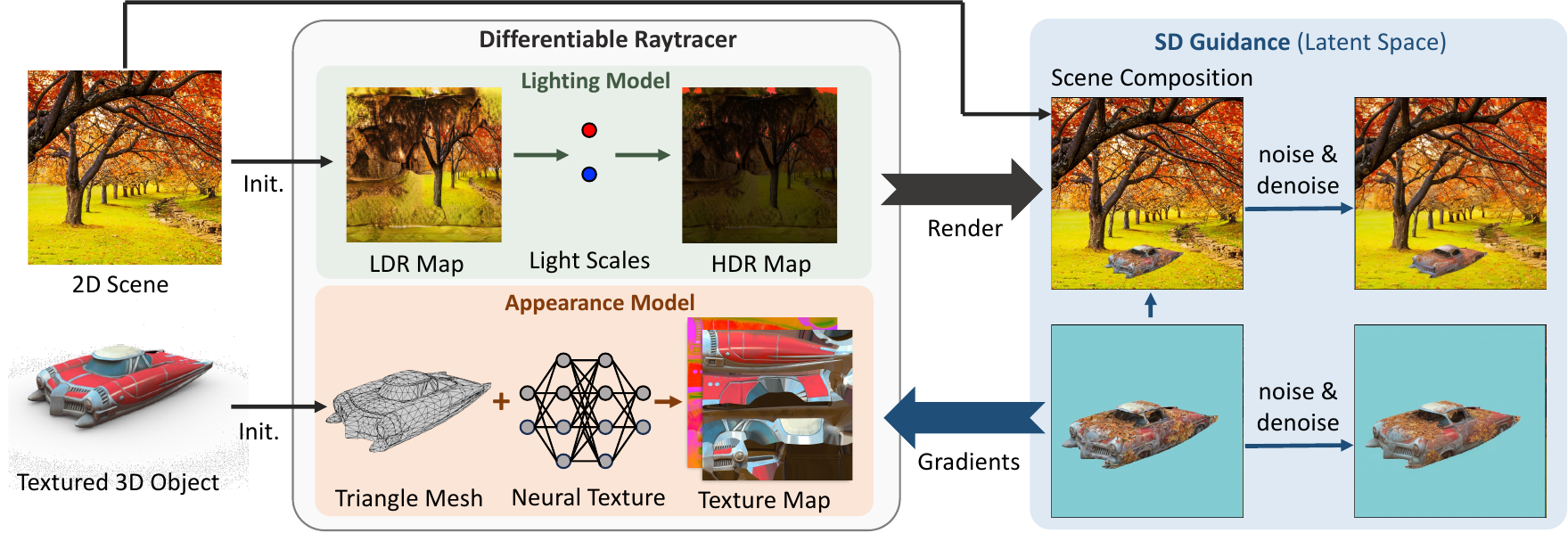}
    \vspace{-0.0cm}
    \caption{\textbf{Framework.} We learn an environment map and a texture map \textit{separately} from the 2D supervision. 
    We initialize (\textit{init.}) the environment map with an LDR map estimated from the 2D scene and learn light multiplying scales bright areas, yielding an HDR map. 
    We employ the PBR material model for texture maps, encoded via MLP with positional encoding. 
    The object is rendered through a differentiable ray tracer and further composed with the scene background, receiving gradients from Stable Diffusion (SD) in the latent space. 
    }
    \label{fig:framework}
\vspace{-0.6cm}
\end{figure*}

\section{Method}
\label{sec:methodology}

The overall structure of our framework can be found in \cref{fig:framework}.
Given a textured 3D object, a 2D image of a scene, and a 2D location where to place the object, we first adapt the texture of the 3D object (\cref{sec:textureadaptation,sec:sceneconditionalguidance}) to align with the scene.
To enable a photorealistic composition of the object and the scene, we further estimate the lighting (\cref{sec:lightingmodel,sec:apparatus,sec:lightconditional}) and apply it during the final rendering.

\subsection{Optimization}
\label{sec:optimization}
To enable end-to-end optimization, we adopt a differentiable renderer to render the object and score it using the priors from diffusion models.

\paragraph{Differentiable Renderer.} We employ the differential renderer Mitsuba3~\cite{mitsuba3} (denoted as $g$), due to its capabilities as a physically-based ray-tracing renderer equipped with versatile rendering options, such as diverse BSDFs, the path tracer integrator, and using an environment map as a light source. 
This enables us to produce photo-realistic renders complete with shadows and reflections, forming the foundation for robust texture adaptation and light estimation.

\paragraph{Diffusion Model as Guidance.} 
To score the rendered image, we make use of a generative diffusion model~\cite{imagen,stablediffusion} $\phi$.
Diffusion models have learned powerful priors of the image formation process in the real world from large-scale training on billion-scale web data.
Similar to prior work~\cite{dreamfusion,sjc}, we supervise the rendered image $\vx_0=g(\vtheta)$ with a score-matching objective to obtain an image gradient
\begin{equation}
\label{eq:vsd}
\nabla_\vtheta\mathcal{L}(\vx_0) \triangleq \mathbb{E}_{t,\epsilon} \left[w_t(\epsilon_\phi(\vx_t) - \hat\epsilon(\vx_t))\frac{\partial\vx_0}{\partial\vtheta}\right],   
\end{equation}
where $\vx_t=\sqrt{\alpha_t}\vx_0 + \sqrt{1-\alpha_t}\epsilon$ is the noisy rendered image, $\epsilon_\phi(\vx_t)$ is an approximated score for a noisy real image, $t \in \mathbb{N}$ indexes the discrete diffusion steps, and $\hat\epsilon$ is a scoring function for $\vx_t$.  Specifically, $\hat\epsilon$ is set as constant $\epsilon$ in Score Distillation Sampling ~\cite{dreamfusion} (SDS), and $\epsilon_\psi$ is an online LoRA-tuned network initialized from $\epsilon_\phi$ in Variational Score Sampling~\cite{prolificdreamer} (VSD).
Notably, we employ their interpolated variant given by $\hat\epsilon=\lambda\epsilon_\psi+(1-\lambda)\epsilon$ with a $\lambda$ annealed from $1$ to $\lambda_e$, helping to counteract the over-texturization often observed in VSD, especially for objects with simple surfaces.
With residual of noises $\epsilon_\phi-\hat\epsilon=(\epsilon_\phi-\epsilon)-\lambda(\epsilon_\psi-\epsilon)$, we highlight its relation with the image gradient obtained in traditional inverse rendering setup $\nabla_\vtheta\mathcal{L}(\vx_0)=\vx_0-\vx_{\mathrm{gt}}$ by reformulating it as
\begin{equation}
\label{eq:relation}
\epsilon_\phi-\hat\epsilon=\sqrt{\frac{\alpha_t}{1-\alpha_t}}[(\vx_0-\hat\vx_\phi)-\lambda(\vx_0-\hat\vx_\psi)],
\end{equation}
where the target used to supervise $\vx_0$ is predicted clean image $\hat\vx_\phi$ from $\epsilon_\phi$ as the positive and $\hat\vx_\psi$ from $\epsilon_\psi$ as the negative. 
The technique enables nominal classifier-free guidance~\cite{classifierfree} 
and thus bypasses mode-seeking behaviors and saturated texture. 
Inspired by~\cite{texture}, we incorporate the inner product of the camera pose and object's surface normal $\langle c,\vn\rangle$ into $w_t$ to penalize faces at small grazing angles.
In practice, we use Stable Diffusion~\cite{stablediffusion} (SD) and absorb the encoder into $g$, optimizing in latent space.

\subsection{Appearance Model} 
\label{sec:textureadaptation}


\paragraph{Textured Mesh.} 
We represent the 3D object as a textured mesh, which is the most common format for 3D assets in the industry and is compatible with many established workflows.
Textured meshes are available in abundance in online datasets and marketplaces~\cite{objaverse,polyheaven}, and recently, one can also generate reasonable quality 3D meshes with text descriptions~\cite{fantasia3d,prolificdreamer,magic3d} or limited viewpoints~\cite{liu2023one,magic123}.

\paragraph{Neural Texture.} 
We represent the appearance of objects using a texture map $\mathcal{T}\in\mathbb{R}^{H_t\times W_t\times5}$, facilitating compatibility of textured results with widely-used rendering engines, such as Blender.
We parameterize the texture using a coordinate neural network that maps 3D coordinates to the texture attributes.
We first map the 3D coordinates of vertices using a rasterizer~\cite{nvidiffrec} based on their UV coordinates into the space of the texture map, resulting in $\mathcal{C}\in\mathbb{R}^{H_t\times W_t\times3}$. 
We then use a 2-layer parameterized MLP without bias and Hash positional encoding $\Phi:\mathcal{C}\rightarrow\mathcal{T}$ to map the coordinates to their texture attributes.
For the final layer, we re-map the output activated after $\mathrm{sigmoid}$ according to pre-defined minimal and maximal values. 
The five channels in $\mathcal{T}$ correspond to the Physically-Based Rendering (PBR) shading model~\cite{shading}, which includes a diffuse term $k_\mathrm{d}\in\mathbb{R}^3$ and an isotropic specular GGX lobe~\cite{ggx} described by $k_\mathrm{rm}=(k_r,k_m) \in\mathbb{R}^2$, with $k_r$ representing roughness and $k_m$ indicating metalness. 
The PBR model, with $k_r$ affecting the scattering of light and $k_m$ influencing the reflective properties, creates realistic light interactions between surfaces.

\begin{figure}[!t]
    \centering
    \includegraphics[width=0.9\linewidth, trim=0 0mm 0 0mm, clip]{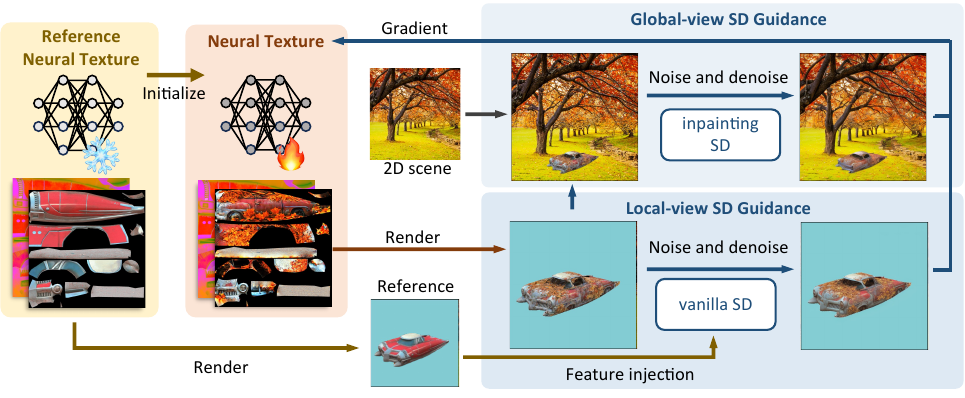}
    \vspace{-0.0cm}
    \caption{\textbf{Pipeline for texture adaptation.} 
    We initialize the neural texture from the reference object and inject the feature of reference renderings to the U-Net of SD. We use both \textit{local-view} and \textit{global-view} guidance.
    }
    \label{fig:texturization}
\vspace{-0.5cm}
\end{figure}

\subsection{Scene-Conditional Guidance}
\label{sec:sceneconditionalguidance}

Achieving a realistic texture adaptation for 3D objects placed into 2D scenes requires three components: \textit{Environmental Influence} involves adjusting the texture to reflect the environmental impact realistically; \textit{Identity Preservation} aims to preserve the objects' unique visual aspects, and \textit{Blending} focuses on guiding the texture adaptation to match the visual characteristic of the scene, ensuring seamless integration with the surroundings and avoiding stark contrasts. 

\paragraph{Environmental Influence.} 
\label{sec:gptprompt}
To show the realistic effects of the scene environment on the object's appearance, we guide the diffusion model with text prompts.
We find that a naive text prompt, combining a simple object description with a scene description, \eg, \textit{a leather sofa in a swamp}, is insufficient to generate realistic textures that reflect the environmental impact on the object, as shown in the appendix.
For instance, we would expect a sofa placed in a swamp to be dirty and possibly show signs of moss or other vegetation growth.
Instead of manually crafting a detailed text prompt for each object and scene combination, we automatize the process by employing a large language model (LLM), specifically GPT-4~\cite{gpt4}, to generate a text prompt for the image diffusion model.
Given the simple description of the object (\eg, \textit{a leather sofa}) and the scene (\eg, \textit{a swamp}), we instruct the LLM first to analyze the potential impact of the scene's environment on the object's appearance and then to distil this information into a text prompt that describes the resulting appearance of the object (\eg, \textit{The leather sofa, partially submerged in the swamp, looks discolored and soggy, its once-polished surface marred by mud, moss, and the murky water of the swamp environment.}).
We then use this automatically generated text prompt to guide the diffusion model in adapting the texture.
The specific prompt design for this automated process is detailed in the appendix.

\paragraph{Identity Preservation.}
\label{sec:rfi}
To prevent the transferred texture from drifting away from its original content, we regularize the denoising diffusion model with reference images, a technique that has been explored for controlled image generation~\cite{pnp,zeropix2pix}. 
Suppose we have a rendered reference image $\vx^r_0=g(\vtheta^r)$ and its noised version $\vx^r_t$ as input forwarded to $\epsilon_\phi$ and $\epsilon_\psi$, for the Multi-Head Self-Attention (MHSA) layer in each Transformer block of $\epsilon_\phi$'s and $\epsilon_\psi$'s U-Nets, we concatenate the key $K^r$ and value $V^r$ of $\vx^r_t$ to the original $K$ and $V$ from $\vx_t$ yielding an image-conditional attention output
\begin{equation}
    A(\vx_t,\cdot,\vx_t^r) = \mathrm{MHSA}(Q,[K;K^r],[V; V^r]),
\end{equation}
where $\cdot$ can either be $\mathrm{text}$ or $\emptyset$ for text conditioning.
The text-unconditional attention $A(\vx_t,\emptyset,\vx_t^r)$ used for classifier-free guidance~\cite{classifierfree} is further modified as $s_\mathrm{c}\cdot A(\vx_t,\emptyset,\emptyset)+(1-s_\mathrm{c})\cdot A(\vx_t,\emptyset,\vx_t^r)$ weighted by a control guidance scale $s_\mathrm{c}$.
With the percentage $p$ of blocks to be injected, the intensity of preservation can be controlled: $s_\mathrm{c}=p=1$ provide the most and $s_\mathrm{c}=p=0$ the least preservation.

\paragraph{Blending.} 
\label{sec:blending}
The two components previously mentioned directly utilize object renderings, referred to as the \textit{local view}, as inputs for vanilla SD. 
To further enhance the adaptation of the object's appearance to its surroundings, we input the scene image, with the object rendered onto it, referred to as the \textit{global view}, into an additional inpainting SD~\cite{stablediffusion}.
We have found that the inpainting model provides superior conditioning on the surrounding environment depicted in the image, resulting in improved blending.
However, the inpainting model can sometimes attempt to undesirably modify the inserted object, removing parts of it by inpainting them with the background.
To counter this, we compose the \textit{local view} with randomly sample solid background colors, rather than the scene.
This prevents the inpainting model from removing parts of the object as the object must look realistic when rendered in front of a random background.
In practice, we utilize a set of \textit{global view}s created by cropping the original \textit{global view} at various scales, ensuring that the crops encompass the entire \textit{local view}.

\begin{figure}[!t]
    \centering
    \includegraphics[width=\linewidth]{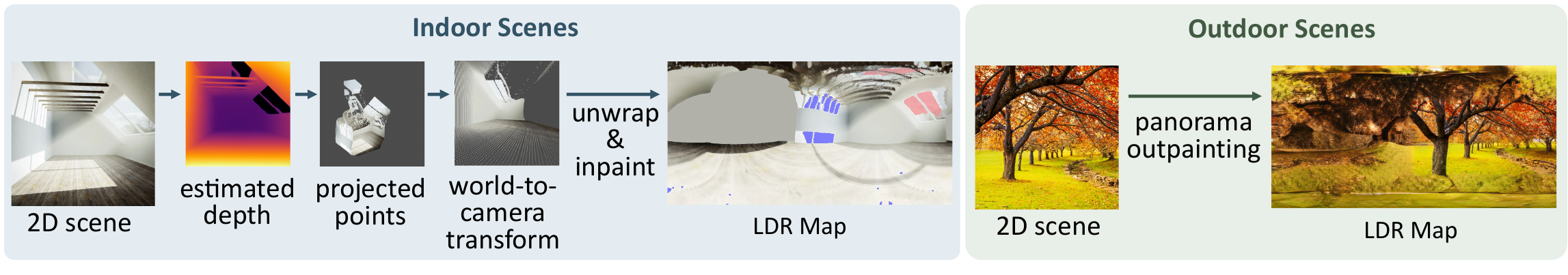}
    \vspace{-0.5cm}
    \caption{\textbf{Pipeline for estimating the LDR map.} We utilize tailored pipelines for indoor and outdoor scenes. Areas masked red and blue correspond to the far light \red{\faSunO} and near light \blue{\faLightbulbO} region.}
    \label{fig:lightingldrpipeline}
\vspace{-0.6cm}
\end{figure}
    

\begin{figure*}
\centering
\includegraphics[width=\textwidth]{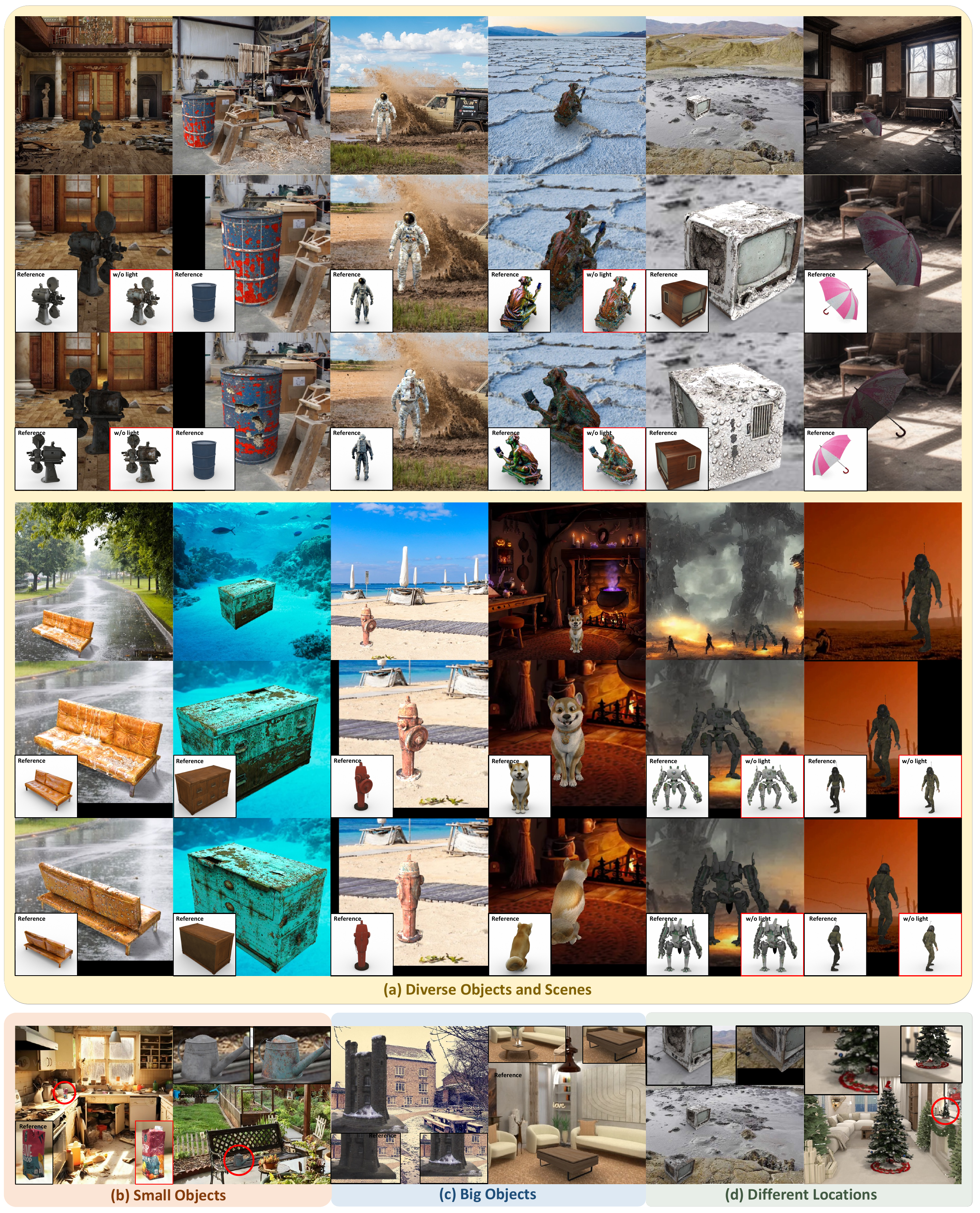}
\vspace{-0.0cm}
\caption{\textbf{Visual Results.} 
\textbf{(a)} We showcase that our method applies to a diverse range of objects and scenes.
The \textit{global view} (top row) the overall composition quality and object-centric \textit{local view} (bottom two rows) for the fidelity of stylized textures are demonstrated. For dim scenes, we additionally render objects without the estimated lighting condition (\textit{w/o light}) for illustrative purposes.
Additionally, we showcase that our method applies to both \textbf{(b)} small and \textbf{(c)} big objects, as well as \textbf{(d)} different placing locations. The texture of the \textit{television}, for example, adjusts the texture to match its surroundings.
}
\label{fig:visualresults}
\vspace{-0.0cm}
\end{figure*}

\subsection{Lighting Model}  
\label{sec:lightingmodel}
To seamlessly integrate the object into the scene, we estimate the lighting conditions, as outlined in \cref{fig:lightingldrpipeline}.
To light the object, we utilize a high dynamic range (HDR) environment map $\mathcal{E}\in\mathbb{R}^{H_e\times W_e\times3}$, which is well-suited for representing natural illumination. 
Since the given 2D scene image only captures a small angle of the full 360-degree environment map, we first use the image as input to estimate a low dynamic range (LDR) environment map.
This process depends on the scene type (\ie, indoor \textit{vs.} outdoor) and is explained in the following paragraphs.
Given the LDR map, we then convert it to an HDR map by estimating a light scale (\ie, a scalar)  for each bright region, which is thresholded from the LDR map directly (\eg, where intensity $\mathcal{I}_{i,j}\geq0.8$).
This scalar is then multiplied by the RGB values to approximate an HDR map.
Note that we use the HDR map only to light the object. It thus does not need to be of high accuracy to make the object look plausible. 

\paragraph{Indoor Scenes.} 
The well-bounded 2D visual geometry of indoor scenes is primarily characterized by enclosed spaces and interconnected planes, facilitating straightforward (rough) 3D reconstruction.
To simulate the physical lighting of indoor scenes, we lift each pixel from the 2D scene to 3D space using its estimated depth $\mathcal{D}$~\cite{midas,omnidata}.
We perform world-to-camera and spherical-to-equirectangular coordinate transformations in succession, unwrapping 3D points into a latitude-longitude formatted LDR map.
In post-processing, we automatically fill holes and remove small isolated regions. 
We fill the unseen region (\ie areas behind the camera) with the average RGB value of the 2D scene.
Two scalars are optimized: one for the far light region (\red{\faSunO} $\triangleq \{\red{i,j} \ | \ \mathcal{D}_{\red{i,j}}\geq \tau_d \land \mathcal{I}_{\red{i,j}}\geq\tau_f$\})  and another for the near light region (\blue{\faLightbulbO} $\triangleq \{\blue{i,j} \ | \ \mathcal{D}_{\blue{i,j}}< \tau_d \land \mathcal{I}_{\blue{i,j}}\geq\tau_n$\}). 

\paragraph{Outdoor Scenes.} 
3D reconstruction of outdoor scenes is challenging due to their inherently unbounded geometric structure. 
Furthermore, in most outdoor scenes, the sun is the dominant light source, and the position change of an object has a negligible effect on the angle shift of the light.
Therefore, instead of painstakingly recovering location-specific viewpoints, we use a single-view to panorama outpainting model~\cite{omnidreamer}, taking its generated output as the initial LDR map.
We optimize one scalar for the far light area with positive elevation (\red{\faSunO} $\triangleq \{\red{i,j} \ | \ \frac{\red{i}}{H_e} < 0.5 \ \lor \ \mathcal{I}_{\red{i,j}}\geq\tau_o\}$), leaving the near light area (\blue{\faLightbulbO} $\triangleq \varnothing$) unoptimized.
While more sophisticated designs, such as a position estimator~\cite{everlight} or sky modeling~\cite{skymodel}, could be incorporated, we find our approach robust enough to provide a sufficient estimation.

\subsection{Light-Capturing Apparatus} 
\label{sec:apparatus}
To estimate light scales, we place 3D objects with a fixed appearance model into the 2D scene and then optimize the lighting with VSD as described in~\cref{eq:vsd} to anchor the lighting impact imposed by the scene.  
Inspired by the traditional inverse rendering setup~\cite{dip} where the environment map can almost be perfectly reconstructed from objects' reflections, we introduce a novel concept incorporating a virtual light-capturing apparatus alongside the object of interest during the optimization process.
We insert a white sphere made of a smooth diffuse material into the scene and guide the lighting optimization with the diffusion model using the text prompt ``\textit{A gigantic diffuse white (spray-painted) sphere (ball)}''.
The white diffuse sphere proves advantageous in stabilizing lighting estimations for scenes with potentially strong light sources, as the reflected intensity from a white diffuse object closely approximates the intensity of the environmental lighting.
In our initial experiments, we also tested spheres with mirrored and matte silver materials but found them less beneficial.

\subsection{Light-Conditional Guidance}
\label{sec:lightconditional}
The absence of light conditions in text prompts makes the generated images only loosely constrained on lighting, which makes it challenging for diffusion models to score the renderings accurately. 
A white sphere looks gray in a darker environment, so light-agnostic prompts can engender an overestimation of the brightness for dim scenes.
Therefore, we append ``\textit{in a dark environment}'' if the average intensity of the background and light areas are below certain thresholds. 
We condition the LoRA-tuned model $\epsilon_\psi$ by concatenating the light scales with the camera extrinsic~\cite{prolificdreamer} as class embeddings for the U-Net. 
This improves light estimation, especially for outdoor scenes and dimly lit scenes.
For scenes with atypical lighting that significantly alters the appearance of objects, we manually append color prompts such as ``\textit{in blue tint}'' for the seabed bathed in blue light, or ``\textit{in red illumination}'' for a nightclub enveloped in red atmospheric light.
\section{Experiments}
\label{sec:experiments}

\begin{figure*}[!t]
\centering
\includegraphics[width=0.98\textwidth]{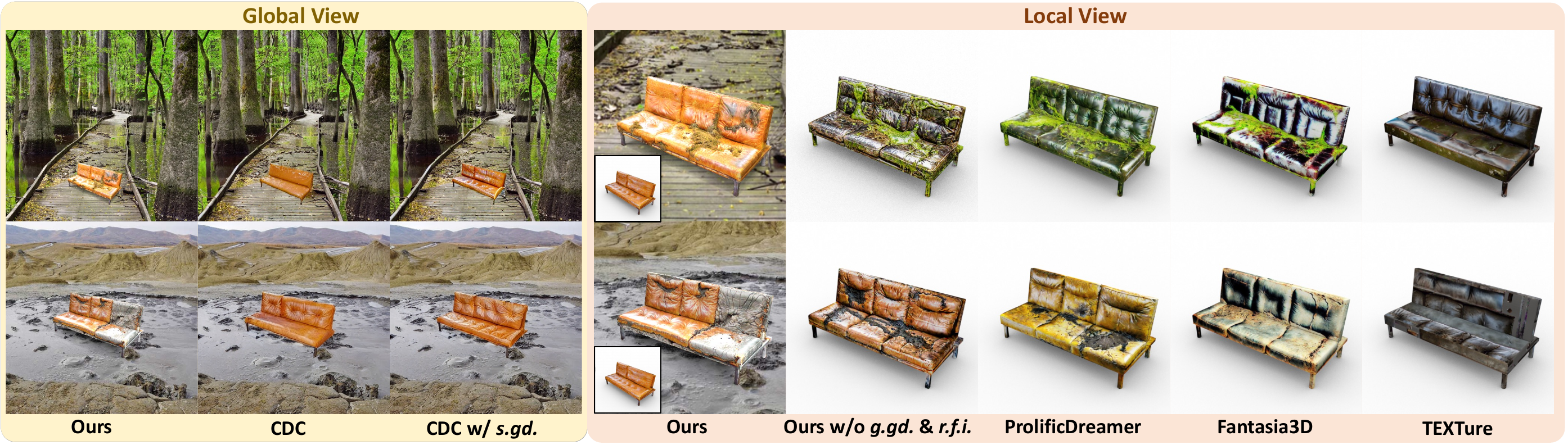}
\vspace{-0.2cm}
\caption{\textbf{Qualitative Comparison.} 
We compare our the generated \textit{global view} with the scene-object compositing method: CDC~\cite{cdc} and the \textit{local view} with mesh texturing methods: Prolific Dreamer~\cite{prolificdreamer}, Fantasia3D~\cite{fantasia3d}, and TEXTure~\cite{texture}. \textit{s.gd.} denotes similarity guidance. \textit{g.gd.} and \textit{r.f.i} denotes global-view guidance and reference feature injection, respectively. Our method achieves both seamless scene-object composition as well as texture adaptation of high fidelity.
See sup. mat. for extensive comparison.
}
\label{fig:comparison}
\vspace{-0.5cm}
\end{figure*}

In this section, we show extensive qualitative and quantitative results of our method in various settings. 
Please see the appendix for many more examples, animations and discussion on limitations and negative impact.

\subsection{Qualitative Evaluation}
\label{sec:results}
We first present visual results and a case study to demonstrate the generalizability of the proposed method.
We then compare our method with others that specifically focus on either scene composition or object stylization.

\begin{figure}[!t]
\centering
\includegraphics[width=\columnwidth]{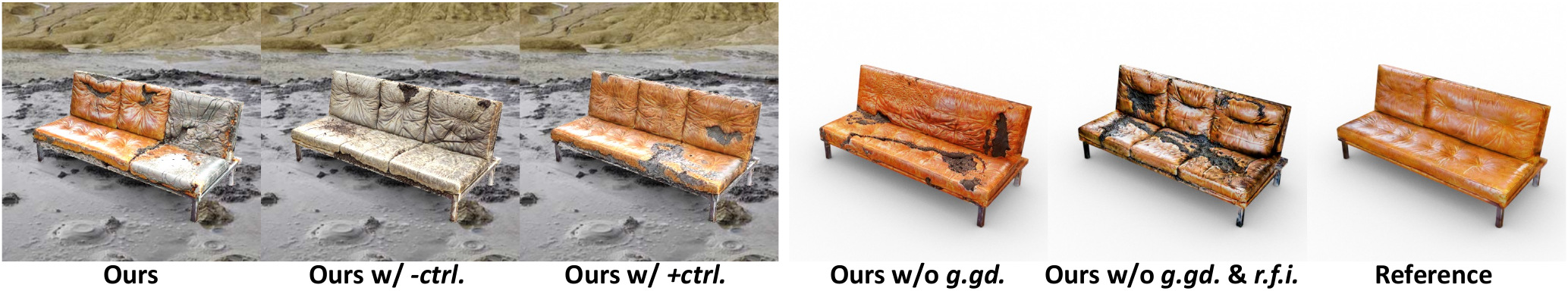}
\vspace{-0.5cm}
\caption{\textbf{Ablation on Texture Adaptation.} We ablate global-view guidance (\textit{g.gd.}) and reference feature injection (\textit{r.f.i}). \textit{-ctrl.} and \textit{+ctrl.}  denotes less or more control in feature injection (smaller or higher $s_\mathrm{f}$ \& $p$) for \textit{Identity Preservation} requirement. }
\label{fig:abl_texture}
\vspace{-0.5cm}
\end{figure}

\paragraph{Main Results.} 
To demonstrate the generalizability of the framework and the impact of different scenes on the inserted objects, we present visual results in~\cref{fig:visualresults}. 
Our method successfully blends the objects into various environments, achieving photorealistic adaptation for both appearance and lighting. This includes scenarios drawn from both real-world and fantasy settings.
For example, the dimly-lit umbrella in~\cref{fig:visualresults} casts a shadow that is perfectly aligned with its handle and the light direction of the scene, substantially increasing the photorealism of the composition; The Belweder TV set is wrapped with gray mud consistent with the scene while its original structural details get maintained.

\paragraph{Comparison on Scene Composition.} 
We showcase the result from CDC~\cite{cdc} using LLM-prompted text prompts for a fair comparison.
As shown in~\cref{fig:comparison}, CDC loses the structural details of the object while providing limited adaption for texture transfer. Moreover, the composed scene appears artificial and synthetic, giving the impression that the sofa is floating above the mud. 
Our experiments reveal that the CDC often leads to object removal, particularly when the re-painting~\cite{repainting} is employed.
Therefore, we further incorporate similarity guidance similar to the classifier guidance~\cite{guideddiffusion} by adding $\sqrt{1-\alpha_t}\cdot\frac{\partial(||\vx^r_0 - \hat\vx_\phi||^2)}{\partial\vx_t}$ to the predicted gradient $\epsilon_\phi(\vx_t)$ each denoising step refered as \textbf{CDC w/ \textit{s.gd.}}, but find it provides limited improvement.

\paragraph{Comparison on Object Stylization.}
We showcase results from mesh texturing method~\cite{prolificdreamer,fantasia3d,texture} using our LLM-prompted text prompts for a fair comparison. As shown in~\cref{fig:comparison}, although achieving realistic texture adaptation, these mesh texturing methods fall short in preserving the original structural details of the object and in utilizing scene-level information. Essentially, These methods only satisfy the \textit{Environmental Influence} requirement via text prompts
and do not meet all the criteria of our intended goal.

\begin{table}[t]
    \centering
    \caption{\textbf{Quantitative Evaluation.} \textbf{(a)} We measure LPIPS for controllability, CLIP$_\mathrm{dir}$ and CLIP$_\mathrm{SI}$ for composability. \textbf{(b)} We present the percentages of users who prefer our method over other methods: CDC~\cite{cdc} and InstructP2P~\cite{instructpix2pix}.}
    \label{tab:quantitative}
    \scriptsize
    \begin{subtable}{0.5\textwidth}
        \centering
        \begin{tabular}{l|ccc}
            Method & LPIPS $\downarrow$ & CLIP$_\mathrm{dir}$ $\uparrow$ &  CLIP$_\mathrm{SI}$ $\uparrow$ \\
            \toprule
            InstructP2P~\cite{instructpix2pix} & 0.0020 & 0.0175 & 0.1852 \\
            Ours (\textit{T.A.}) & \textbf{0.0014} & 0.0201 & 0.2006 \\
            \midrule
            CDC~\cite{cdc} & 0.0053 & 0.0042 & 0.1996 \\
            Ours (\textit{T.A.} + \textit{L.E.}) & 0.0015 & \textbf{0.0206} & \textbf{0.2013} \\
            \bottomrule
        \end{tabular}
        \subcaption{Visual Metrics.}
        \label{subtab:left}
    \end{subtable}%
    \hspace{0.3cm}
    \begin{subtable}{0.45\textwidth}
        \centering
        \setlength{\tabcolsep}{2pt}
        \begin{tabular}{l|ccc}
            & Ours & Draw & Other \\
            \toprule
            Ours vs. CDC~\cite{cdc} & \textbf{71\%} & 20\% & 9\% \\
            \midrule
            Ours vs. InstructP2P~\cite{instructpix2pix} & \textbf{74\%} & 6\% & 20\% \\
            \bottomrule
        \end{tabular}
        \subcaption{User study for human preference.}
        \label{subtab:right}
    \end{subtable}
\vspace{-1.0cm}
\end{table}

\subsection{Quantitative Evaluation}
We collected 34 composites between diverse objects and scenes, each with 3 random views, yielding a total of 102 images.
Besides CDC~\cite{cdc}, we compare with InstructPix2Pix~\cite{instructpix2pix} by directly copy-pasting the object with adapted texture onto the scene.
We measure LPIPS for controllability, CLIP$_\mathrm{dir}$ and CLIP$_\mathrm{SI}$ for composability, similar to CDC, and report the results in~\cref{tab:quantitative}.
Compared with InstructPix2Pix our method for texture adaptation (\textit{T.A}) alone excels in controllability and composability.
Light estimation (\textit{L.E.}) further improves results. 

We also conducted a \textit{user study} in ~\cref{tab:quantitative} with 42 participants to compare the realism of our compositing under the condition that the object identity must be preserved.
Our method is preferred over prior work in over 70\% of the cases.

\begin{figure}[!t]
    \centering
\scalebox{0.8}{\includegraphics[width=\columnwidth]{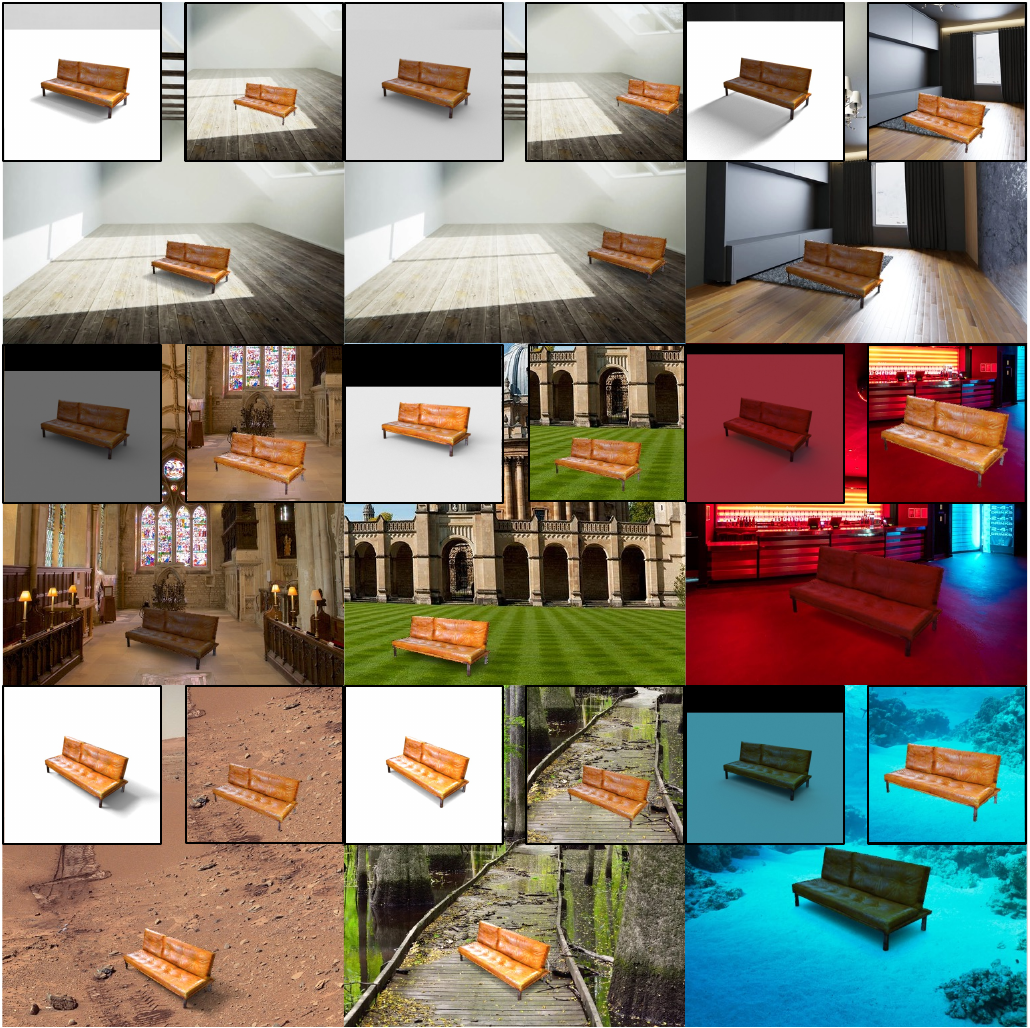}}
\vspace{-0.0cm}
\caption{\textbf{Ablation on Light Estimation.} We evaluate the effectiveness of light estimation by demonstrating the lighting effects on a sofa in various indoor and outdoor scenes. 
Overlaid in the left corners are objects rendered with the estimated light against a white background.
In the right corners are objects before the light estimation.
}
\label{fig:abl_light}
\vspace{-0.5cm}
\end{figure}

\subsection{Ablation Study} 
\label{sec:ablation}

\paragraph{Identity Preservation.} 
To illustrate how the \textit{Identity Preservation} requirement is satisfied by reference feature injection (\textit{r.f.i.}), we showcase the leather sofa in the swamp with less and more control (\cref{fig:abl_texture}). The setup with less control drifts to a visually different sofa with too much mud wrapped around while more control generates a less muddy and adapted sofa compared to the reference image. Both adapted textures are less visually plausible.
Additionally, when comparing \textbf{Ours w/o \textit{g.gd.} \& \textit{r.f.i.}} to the reference image, it is evident that the identity of the sofa is lost.

\paragraph{Global-View Guidance.} 
To showcase the necessity of global-view guidance we conduct texture adaptation without it as \textbf{Ours w/o \textit{g.gd.}} in ~\cref{fig:abl_texture}.
The adapted texture without seeing the full scene fails to adapt the color of the mud to that of the scene image, contrasting starkly with the scene.
Experiments ablating the optimal global-view guidance set-up (\eg, solid-color background augmentation, usage of inpainting model, \etc) are detailed in the appendix.

\paragraph{Light Estimation.}
In \cref{fig:teaser}, the combination of accurately estimated lighting and shadows significantly enhances the photorealism of the composed image (compare the image on the left, which includes lighting, to those in the top right corner, which are without lighting). To further test the validity of the estimated light, we render images omitting the texture adaptation step.
\cref{fig:abl_light} demonstrates that the estimated light considers multiple factors, including its direction relative to the specific locations of objects in indoor scenes and its impact on the appearance of objects under atmospheric colored lighting conditions.
Importantly, the estimated HDR maps, paired with the proposed scene composition, facilitate the realistic integration of shadows.

\section{Conclusion}
\label{sec:conclusion}

In this paper, 
we are motivated by a practical goal---\textit{what if a 3D object is placed into a 2D scene?}---and propose a novel framework that allows 
(1) stylizing the object with an adapted texture that aligns with the given scene; and
(2) achieving a photorealistic scene composition with the aid of estimated light of the environment.
This is enabled by physically-based differentiable ray tracing and diffusion models used as guidance.
Our method allows for automatic adaptation and blending of existing objects into a variety of scenes, making it useful for both 2D and 3D downstream applications, such as visual media and video games.
\appendix

\begin{figure*}[th]
\centering
\includegraphics[width=1.0\textwidth]{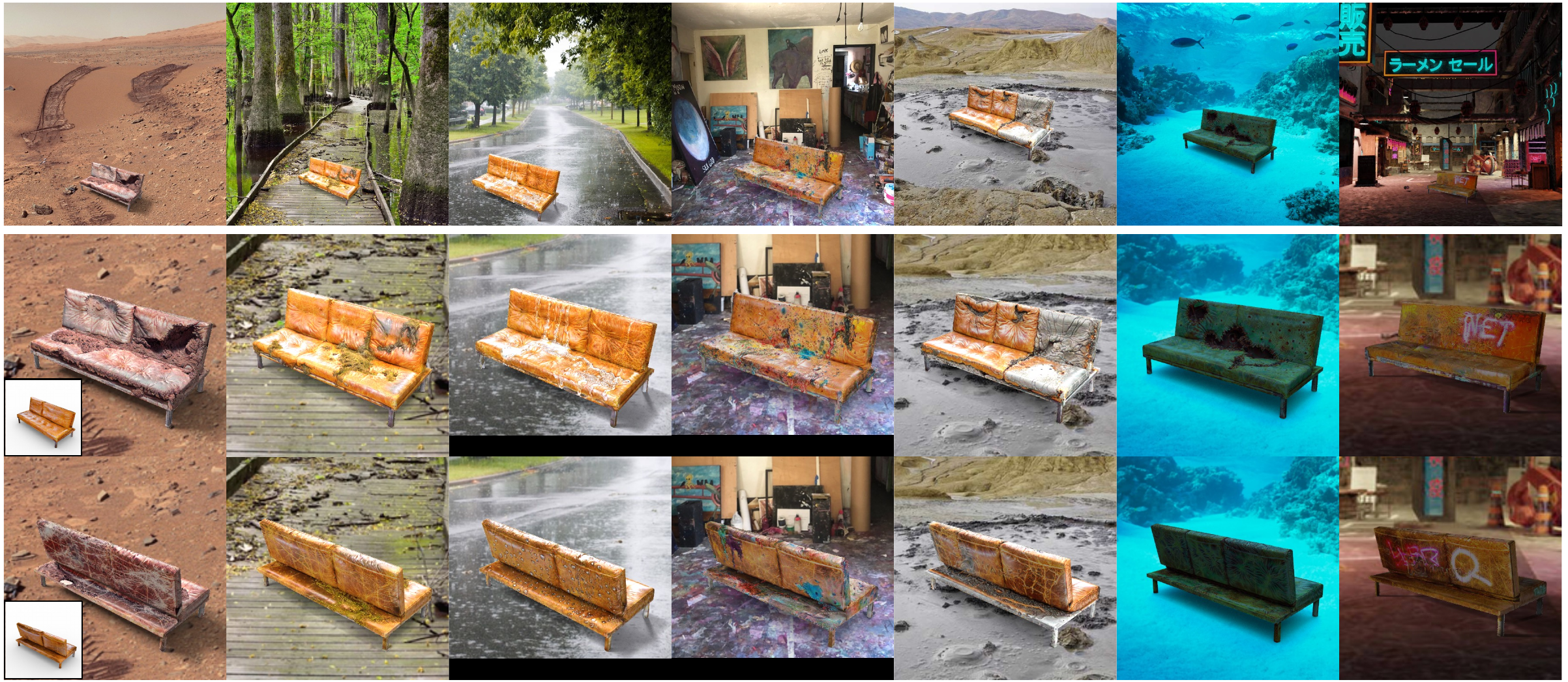}
\vspace{-0.0cm}
\caption{\textbf{Case Study.} 
We showcase the \textit{global view} (top row) to demonstrate the overall composition quality and object-centric \textit{local view} (bottom two rows) for the fidelity of stylized textures for a wide range of different scenes using the same original object: a leather sofa.
}
\label{fig:casestudy}
\vspace{-0.0cm}
\end{figure*}

\begin{figure}[!t]
\centering
\includegraphics[width=0.8\columnwidth]{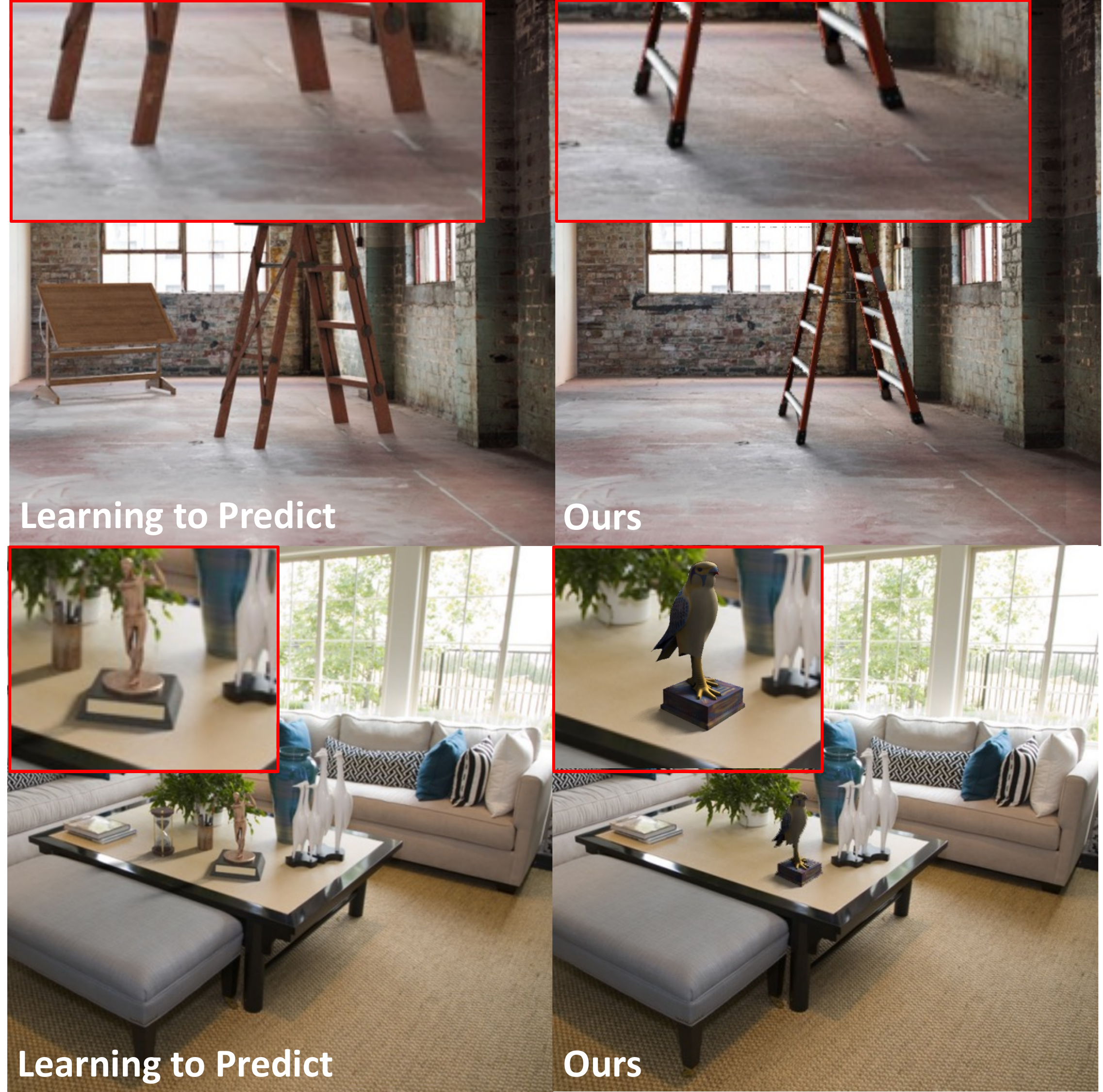}
\vspace{-0.0cm}
\caption{\textbf{Qualitative Comparison on Light Estimation.} 
We compare with a light-estimation method~\cite{indoor1}. 
Our method yields comparable results with shadow cast consistent with the light source in the image.
}
\label{fig:comparison_light}
\vspace{-0.5cm}
\end{figure}

\section{Additional Visual Results}
\label{app:results}

We encourage readers to explore the submitted zip file, which contains numerous visual results, animated results, and videos demonstrating the training dynamics.
Additionally, we conduct a case study in~\cref{fig:casestudy}, where we specifically place a sofa into a diverse array of scenes. 

\paragraph{Light Estimation.} We qualitatively compare with a light estimation method, Learning to Predict~\cite{indoor1}, that is trained on real HDR maps.
Our method yields comparable results with shadow casts consistent with the light source in the image, as demonstrated in~\cref{fig:comparison_light}.


\section{Additional Ablation Study}
\label{app:abaltion}

\begin{table*}[t]
\centering
\setlength{\tabcolsep}{0.5mm}{
\scalebox{0.77}{
\begin{tabular}{lc>{\centering\arraybackslash}m{2.5cm}cccc}
 & \multicolumn{3}{c}{Global-View Guidance} & \multicolumn{3}{c}{Local-View Guidance}  \\
\cmidrule(lr){2-4}\cmidrule(lr){5-7}
& Background & Model  & Prompt & Background &  Model  &  Prompt \\
\toprule
\multicolumn{7}{l}{\textcolor{gray!80}{\textit{Scene-Agnostic}}} \\
\textcolor{gray!80}{$\mathrm{(a)}$} & \textcolor{gray!80}{-} & \textcolor{gray!80}{-} & \textcolor{gray!80}{-} & \textcolor{gray!80}{solid color} & \textcolor{gray!80}{vanilla SD} & \textcolor{gray!80}{object} \\
\textcolor{gray!80}{$\mathrm{(b)}$} & \textcolor{gray!80}{-} & \textcolor{gray!80}{-} & \textcolor{gray!80}{-} & \textcolor{gray!80}{solid color} & \textcolor{gray!80}{InstructPix2Pix~\cite{instructpix2pix}}  & \textcolor{gray!80}{manually instructed$^*$} \\
\textcolor{gray!80}{$\mathrm{(c)}$} & \textcolor{gray!80}{-} & \textcolor{gray!80}{-} & \textcolor{gray!80}{-} & \textcolor{gray!80}{solid color} & \textcolor{gray!80}{vanilla SD w/ \textit{r.f.i.}}  & \textcolor{gray!80}{manually combined$^*$} \\
\midrule
\multicolumn{7}{l}{\textit{+ Environmental Influence}} \\
$\mathrm{(d)}$ & - & - & - & solid color & vanilla SD & manually combined \\
$\mathrm{(e)}$ & - & - & - & solid color & vanilla SD & GPT-prompted \\
\midrule
\multicolumn{7}{l}{\textit{+ Environmental Influence + Identity Preservation}} \\
$\mathrm{(f)}$ & - & - & - & solid color & InstructPix2Pix~\cite{instructpix2pix}  & manually instructed \\
$\mathrm{(g)}$ & - & - & - & solid color & InstructPix2Pix~\cite{instructpix2pix}  & GPT-prompted \\
$\mathrm{(h)}$ & - & - & - & solid color & vanilla SD w/ \textit{r.f.i.}  & GPT-prompted \\
\midrule
\multicolumn{7}{l}{\textit{+ Environmental Influence + Identity Preservation + Blending}} \\
$\mathrm{(i)}$ & - & - & - & scene  & vanilla SD w/ \textit{r.f.i.} & GPT-prompted \\
$\mathrm{(j)}$ & scene & vanilla SD & manully combined & solid color & vanilla SD w/ \textit{r.f.i.}  & object \\
$\mathrm{(k)}$ & scene & vanilla SD & GPT-prompted &  solid color & vanilla SD w/ \textit{r.f.i.}  & object \\
$\mathrm{(l)}$& scene & vanilla SD & GPT-prompted  & scene  & vanilla SD w/ \textit{r.f.i.}& object \\
$\mathrm{(m)}$ & scene & inpainting SD & GPT-prompted  & scene & vanilla SD w/ \textit{r.f.i.} & object \\
$\mathrm{(n)}$ & scene & inpainting SD & GPT-prompted & scene & vanilla SD w/ \textit{r.f.i.}  & GPT-prompted \\
\rowcolor{cyan!50}
$\mathrm{(o)}$ & scene & inpainting SD & GPT-prompted  & solid color & vanilla SD w/ \textit{r.f.i.} & GPT-prompted \\
\bottomrule
\end{tabular}}}
\vspace{0.2cm}
\caption{\textbf{Ablative Study on Meta Setup for Texture Adaptation.} 
$\mathrm{(a)}$ is \textit{scene-agnostic texture generation} setup as described in~\cref{app:sceneagnosticimplementation}. 
$\mathrm{(c)}$ is \textit{scene-agnostic texture editing} setup as described in~\cref{app:sceneagnosticeditingimplementation}.
$\mathrm{(o)}$ is the complete \textit{scene-conditional texture adaptation} setup as described in~\cref{app:textureimplementationdetails}.
$^*$ denotes for scene-agnostic text prompts considering the given instruction. For example, with the object prompt being \textit{a leather sofa}, the manually instructed prompt can be ``\textit{add the dusting to the sofa}'' and the manually combined prompt can be ``\textit{a leather sofa with dust}''.
}
\vspace{-0.5cm}
\label{tab:texturemetasetup}
\end{table*}

\begin{figure*}[th]
\centering
\includegraphics[width=0.9\textwidth]{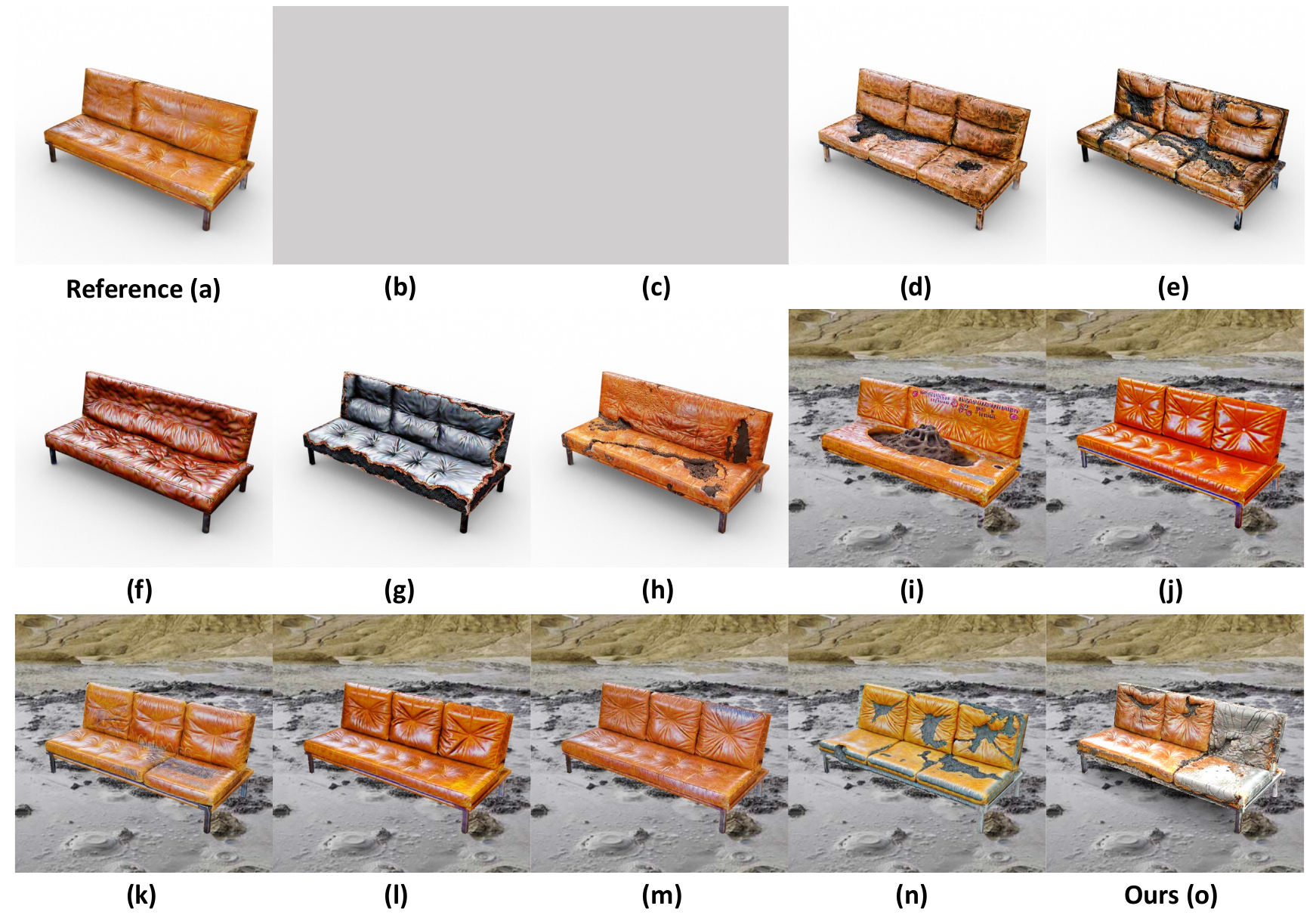}
\caption{\textbf{Additional Ablation on Texture Adaptation.} 
We study the impact of different modules on adapted textures.
$\mathrm{(a)}-\mathrm{(o)}$ correspond to the same settings as in~\cref{tab:texturemetasetup}.
We exclude $\mathrm{(b)}-\mathrm{(c)}$ from this study as they do not adapt an existing texture. 
$\mathrm{(d)}$ and $\mathrm{(e)}$ utilized scene-conditional text prompts to meet the \textit{Environmental Influence} requirement.
$\mathrm{(f)}-\mathrm{(h)}$ studies how \textit{Identity Preservation} can be better achieved. $\mathrm{(i)}-\mathrm{(o)}$ studies how to leverage scene background during adaptation, aiming for desirable \textit{Blending} with objects' placed context.
}
\vspace{-0.5cm}
\label{fig:texturemeta}
\end{figure*}

\subsection{Texture Adaptation}
\label{app:texturemetasetup}

In this section, 
we conduct ablative studies on the meta-setup for texture adaptation \textit{without applying the estimated light} with \textit{sofa} case for a direct comparison
We summarize the different setups in \cref{tab:texturemetasetup} and present the results in \cref{fig:texturemeta}.
We first study the effectiveness of GPT prompts ($\mathrm{(d)}-\mathrm{(e)}$) and compare the proposed reference feature injection with employing InstructPix2Pix~\cite{instructpix2pix} as guidance ($\mathrm{(f)}-\mathrm{(h)}$) similar to \textit{scene-agnostic} setup detailed in~\cref{app:sceneagnosticimplementation} with variations in the choice of model and prompts.
Further, studies investigating the optimal setup of how to incorporate global-view guidance ($\mathrm{(i)}-\mathrm{(o)}$) are conducted. 


\paragraph{Manual \textit{vs.} GPT-Prompted Text Prompts.} 
\label{app:manualvsgptprompt}
We ablate the necessity of GPT-prompted text prompts by comparing $\mathrm{(d)}$ to $\mathrm{(e)}$ and $\mathrm{(f)}$ to $\mathrm{(g)}$. Compared with the manually designed prompts, the GPT-prompted ones lead to more expressive and versatile textures, vividly showcasing the potential \textit{Environmental Influence}s the scene can exert on the objects. 

\paragraph{Reference Feature Injection \textit{vs.} InstructPix2Pix~\cite{instructpix2pix}} 
\label{app:instructpix2pix}
We compare the capability balancing transfer and control with InstructPix2Pix~\cite{instructpix2pix}, validating the effectiveness of the proposed reference feature injection module. Comparing $\mathrm{(g)}$ to $\mathrm{(h)}$, InstructPix2Pix drastically shifts the structural details and identity of the original sofa. The adapted texture from InstructPix2Pix also appears to be less realistic.

\paragraph{Vanilla \textit{vs.} Inpainting Model.}
\label{app:globalviewguidance}
Two factors can be considered when incorporating scene-level guidance: the scene-conditional text prompts and the scene-conditional background. We find that it is better to bind the scene-composed image with an inpainting SD to account for the \textit{Blending} requirement. 
Comparing $\mathrm{(l)}$ to $\mathrm{(m)}$, the inpainting SD encourages the generated textures to be more related to the scene content.
We observe that it is crucial to use GPT-prompted prompts for the vanilla SD. Comparing $\mathrm{(k)}$ to $\mathrm{(o)}$, not using GPT-prompted prompts leads to a much more expressive and adapted texture.

\paragraph{Scene \textit{vs.} Solid-Color Background Composition.}
\label{app:solidbackground}
The input to vanilla SD is object renderings composed with a solid background to ensure objects not camouflaging into the scenes. Comparing $\mathrm{(n)}$ to $\mathrm{(o)}$, a solid background helps to enhance the clarity of the object's boundary pixels. In $\mathrm{(i)}$ where only vanilla SD is used for scene-composed input, the legs of the sofa are entirely blended into the muddy ground, rendering it unrecognizable and inappropriate for downstream usage. 

\begin{figure}
\centering
\begin{minipage}[c]{0.47\textwidth}
\includegraphics[width=\textwidth]{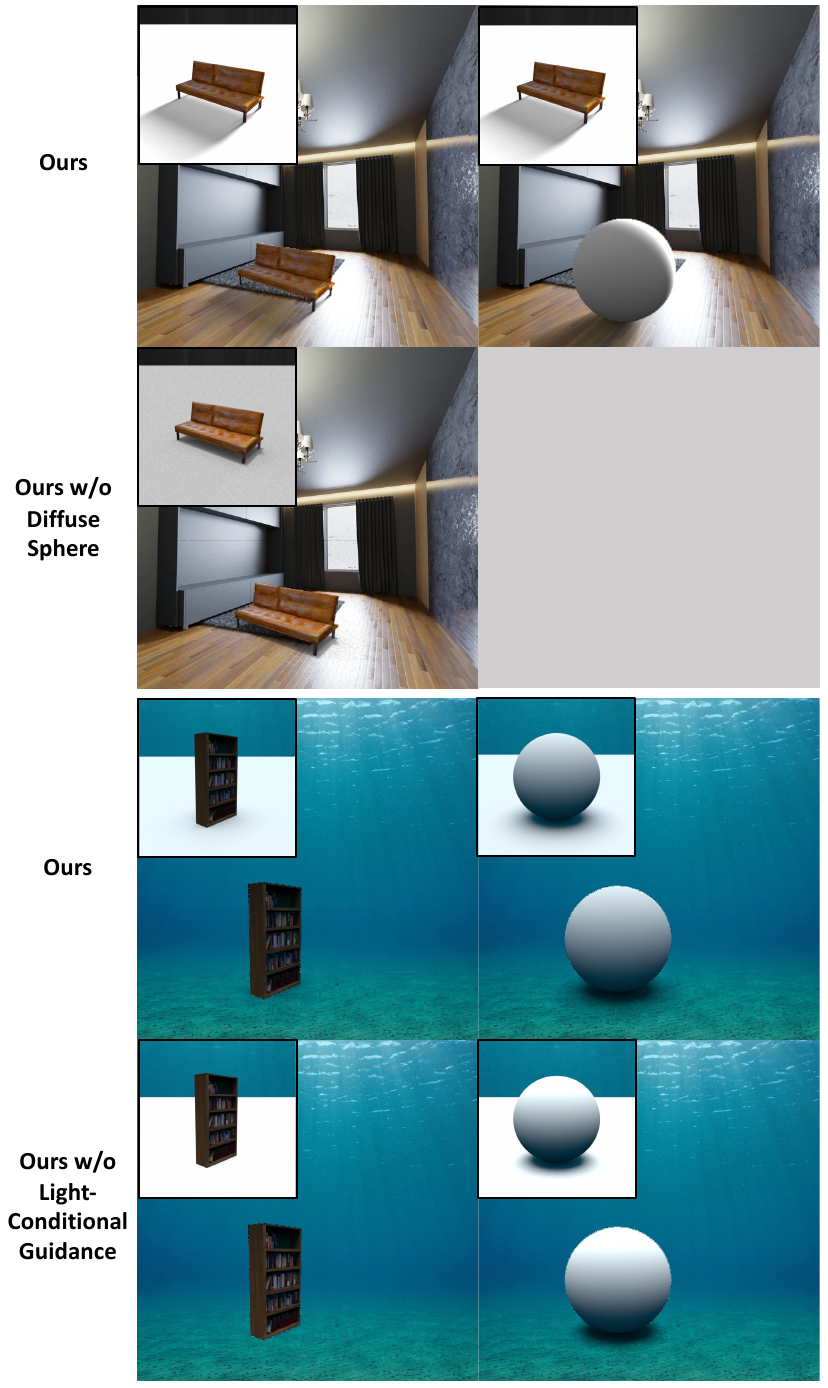}
\caption{\textbf{Additional Ablation on Light Estimation.} 
\textbf{Ours} denotes the complete setup.
We ablate the necessity of light-conditional guidance (first row \textit{vs.} second row) and the white diffusive sphere (third row \textit{vs.} last row). 
Without the diffuse sphere (top) the intensity of the dominant light source from the window is undesirably estimated. 
Without the light-conditional guidance (bottom), the objects appear to be overly bright and do not fit in the dark blue atmosphere of the seabed.
\label{fig:abl_light_ball}
}
\end{minipage}%
\hspace{0.1cm}
\begin{minipage}[c]{0.47\textwidth}
\includegraphics[width=\textwidth]{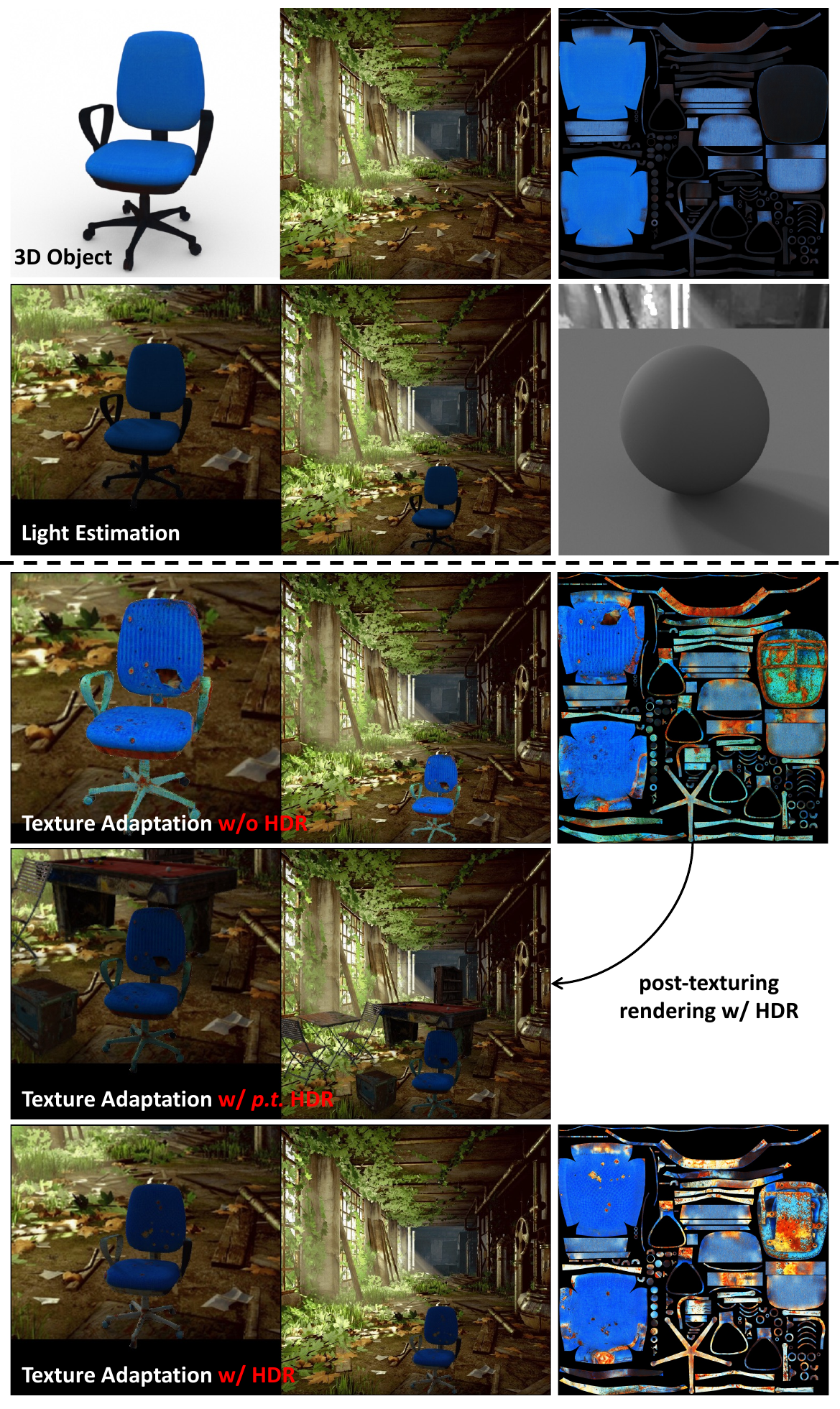}
\caption{\textbf{Texture Adaptation with or without HDR map.} During texture adaptation, we use default ambient light during texture adaptation (w/o HDR, third row) and further apply the estimated HDR map at the post-texturing (\textit{p.t.}) rendering stage (w/ \textit{p.t.} HDR, fourth row).
Alternatively, one can initialize the HDR map (w/ HDR, last row) optimized from the light estimation stage (second row) to decouple potential environmental lighting effects from appearance effects. 
The text prompt we use is ``\textit{The wheeled office chair, upholstery gone, structure rusted, wheels seized, enmeshed in vegetation, forgotten relic of a bygone era.}''}
\label{fig:abl_light_more}
\end{minipage}
\end{figure}


\subsection{Light Estimation}
In this section, we conduct ablation studies on setup for light estimation, justifying the usage of the light-conditional guidance and light-capturing apparatus in stabilizing the estimation.

\paragraph{Light Capturing Apparatus.} We ablate the necessity of a diffuse ball. As shown in the first two rows of~\cref{fig:abl_light_ball}, adding a white diffuse sphere helps to locate the dominant light and optimize its range into a reasonable range.

\paragraph{Light-Conditional Guidance.} 
As mentioned before, adding the dark prompt and color prompt helps optimize for dark scenes and scenes with atypical lighting. As shown in the last two rows of~\cref{fig:abl_light_ball}, light estimation without light-conditional guidance appears to be too bright for the \textit{seabed}.

\begin{figure}[!h]
\centering
\includegraphics[width=\linewidth]{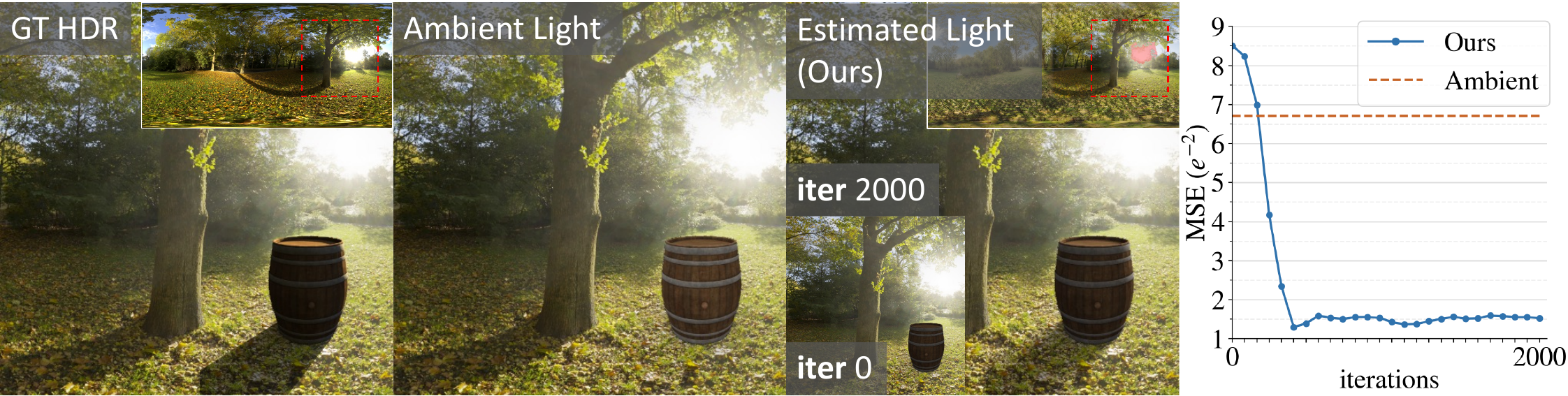}
\vspace{-0.5cm}
\caption{
\textbf{Effectiveness of light estimation}.
}
\vspace{-0.8cm}
\label{fig:abl_light_loss}
\end{figure}

\paragraph{Effectiveness of Light Estimation.}
We visualize the MSE between the object rendered with estimated light throughout training and the one with the GT HDR map in~\cref{fig:abl_light_loss}.
As suggested by the decreasing trend, estimating light scales is sufficient in approximating HDR and thus obtaining decent compositing results.

\subsection{Composition}
\label{app:texturewithlight}

We ablate in this section whether to apply the estimated light condition during the texture adaptation step or only use it in the final rendering for scene composition.
We observe that the integration of an HDR map in the texture adaptation step yields less versatility in the generated texture. 
Nonetheless, in scenarios where the lighting is atypical, such as poorly lit scenes or environments bathed in colored illumination, it is necessary to utilize the HDR map to ensure that the environmental lighting does not unduly influence the appearance of objects.


\paragraph{Texture Adaptation without HDR Map. } As shown in~\cref{fig:abl_light_more}, applying the estimated HDR map during texture adaptation leads to comparatively less expressive adaptation and visual transfer. If the object is poorly lit as the input to the diffusion model, it is hard for the model to derive a clear denoised supervisory signal.

\begin{figure*}
\centering
\includegraphics[width=\textwidth]{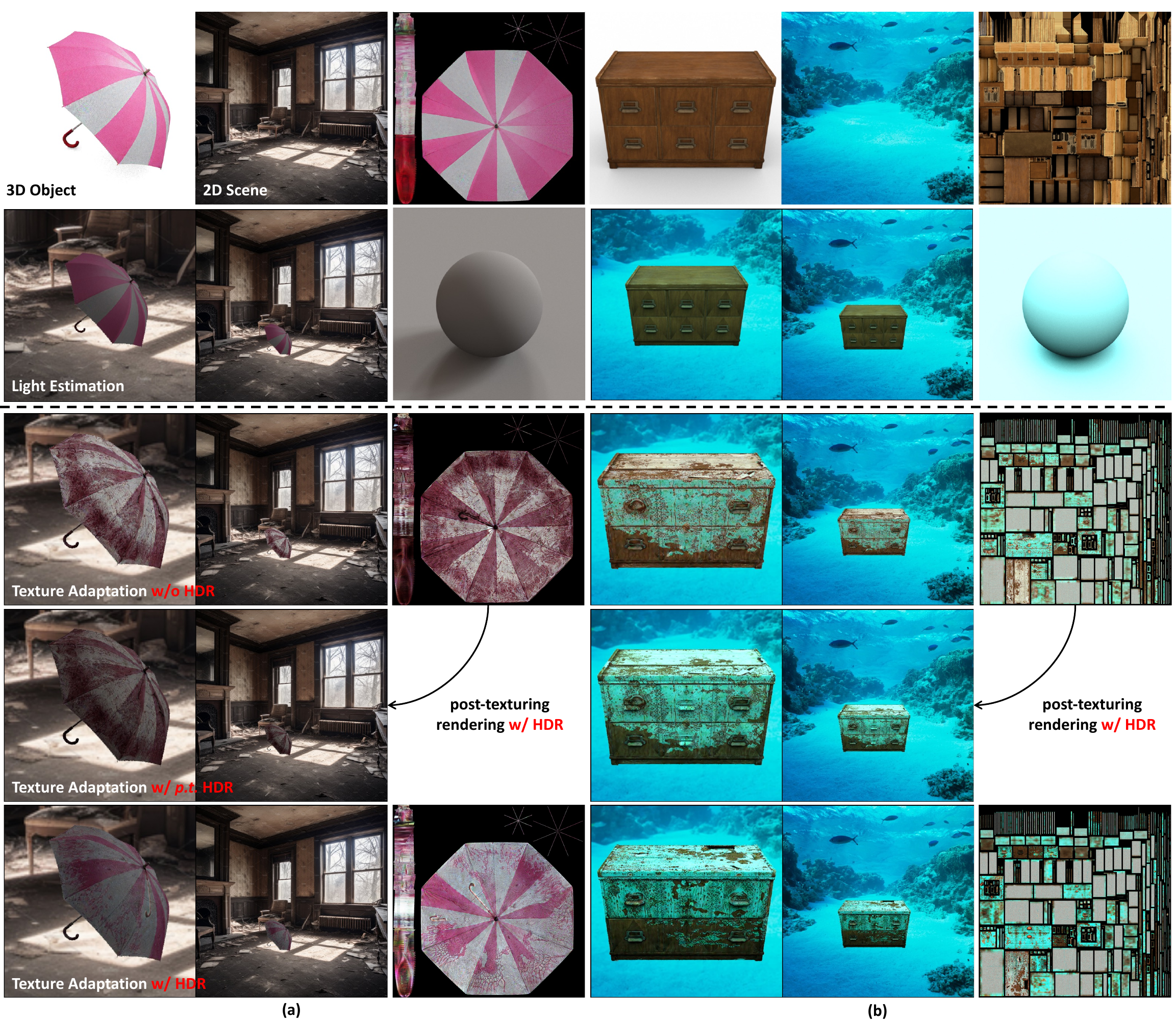}
\caption{\textbf{Texture Adaptation with HDR map in atypical lighting.} During texture adaptation in atypical lighting, we initialize the HDR map (w/ HDR, last row) optimized from the light estimation stage (second row) to decouple potential environmental lighting effects from appearance effects. Alternatively, one can use default ambient light during texture adaptation (w/o HDR, third row) and further apply the estimated HDR map at the post-texturing (\textit{p.t.}) rendering stage (w/ \textit{p.t.} HDR, fourth row). 
Two cases are considered with text prompts (a) ``\textit{Umbrella, dusty, partly opened, shadowed, cobwebs draping, amidst decaying furniture;}'' and (b) ``\textit{Vintage wooden drawer, swollen, colors faded, surfaces colonized by marine life, partly buried in sandy ocean floor, corroding metal fixtures}.''}
\label{fig:texturewhdr}
\end{figure*}

\paragraph{Texture Adaptation with HDR Map in Atypical Lighting. } 
As shown in~\cref{fig:texturewhdr}, omitting the estimated HDR map results in the unintended spill of lighting effects into the object's appearance. Applying the HDR map at the post-texturing stage adversely impacts the photorealism, contracting the optimized texture.



\begin{figure*}
\centering
\includegraphics[height=0.8\textheight]{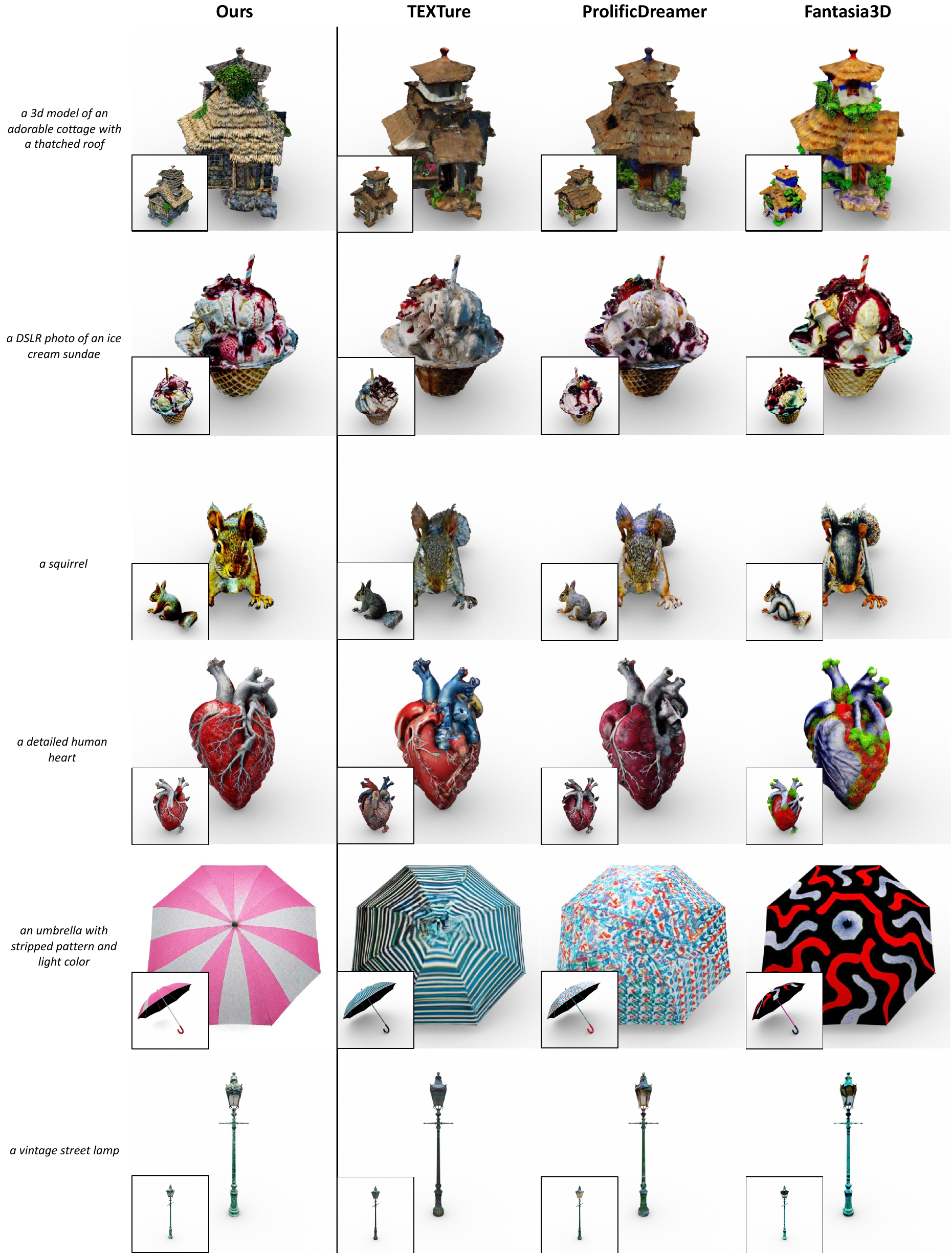}
\caption{\textbf{Scene-Agnostic Texture Generation.} 
We compare our framework with mesh texturing methods: Prolific Dreamer~\cite{prolificdreamer}, Fantasia3D~\cite{fantasia3d}, and TEXTure~\cite{texture}.
We consider textless meshes generated from text prompts or downloaded online.
}
\label{fig:sceneagnostic}
\end{figure*}

\section{Implementation Details}
\label{app:implementationdetails}

\paragraph{Camera Sampling.} 
We sample $24$ views with an azimuth uniformly sampled across $360^\circ$ and a user-specified elevation $e$.
We sample a FOV multiplier $\lambda_\mathrm{FOV}$ such that camera FOV is $\mathrm{tanh}(\frac{r\cdot\lambda_\mathrm{FOV}}{d_\mathrm{cam}})$ where $r=0.5$ is the maximal range of normalized vertices and $d_\mathrm{cam}=2.0$ is the camera distance. A rendering resolution of $512$ and samples per pixel of $128$ are used during training, while samples per pixel of $1024$ are used for evaluation.

\paragraph{Scene Composition.} 
To insert the object into the scene image, we directly alpha-blend the 2D rendering of the object with the scene image according to the user-provided position and size. 
To allow for object shadows, we model an infinitely expansive floor as a plane where the 3D object is placed.
Isolating the shadow from its white background can be done by simple thresholding based on the average intensity of the rendered floor, resulting in two regions: shadowed region $R_{\mathrm{s}}$ and lit region $R_{\mathrm{l}}$. 
We scale the intensity of pixels in $R_{\mathrm{s}}$ with a scalar
$\frac{\mathrm{norm}([1,1,1])}{\mathbb{E}(\{\mathcal{I}_{i,j} \ | \ i,j \in R_{\mathrm{l}})}$ and further convert the scaled intensity into transparency. 

\paragraph{Textured Mesh with Neural Texture.} 
\label{app:texturedmesh}
We consider two sources for the textured mesh with its neural texture, with which the MLP weights of the appearance model can be initialized:
\begin{itemize}
    \item For meshes textured with text prompts and SD guidance, we directly resume the MLP weights. See~\cref{app:sceneagnostic} for its implementation details and a comparison with other scene-agnostic texturing methods.
    \item For meshes downloaded online that come with an existing texture, we first convert them into the neural texture by optimizing the neural network's parameters using inverse rendering to match the original texture, using a learning rate of $0.02$ annealed to $0.001$ for a total of $1000$ iterations on 1 GPU.
\end{itemize}
We bake the texture map and re-map the UVs if multiple materials exist. We re-mesh the mesh and interpolate the UVs if there are large faces. 

\paragraph{Texture Adaptation.}
\label{app:textureimplementationdetails}
For PBR shading model in Mitsuba3~\cite{mitsuba3} API, we opt for the $\mathrm{principled}$ BSDF with $k_d$ as $\mathrm{base\_color}$, $k_r\in\mathbb{R}^1$ as $\mathrm{roughness}$, and $k_m\in\mathbb{R}^1$ as $\mathrm{metallic}$. We sample $\lambda_\mathrm{FOV}\in[1.0, 1.21]$, avoiding degeneration on fidelity due to rendering resolution. We use a fixed ambient light.

As summarized in~\cref{tab:texturemetasetup}, we use $\mathrm{(m)}$ as the complete setup.
Specifically, we use a vanilla SD ($\mathrm{stabilityai/stable{\text-}diffusion{\text-}2{\text-}base}$)
for local-view guidance with \textit{local view} (\ie, the object rendering composed with solid color) as input
and an inpainting SD ($\mathrm{stabilityai/stable{\text-}diffusion{\text-}2{\text-}inpainting}$)
for global-view guidance with scene-composed \textit{global view}s as input. 
The local-view and global-view guidance contribute equally to the final loss.
A $\lambda=1.0$, a control guidance $s_\mathrm{c}=0.0$, a percentage of $p=1.0$ for feature injection, a classifier-free guidance~\cite{classifierfree} $s_\mathrm{cf}=7.5$, and noise levels with $t\in[500, 990]$ are used by default. We find that annealing noise levels from high to low is beneficial in achieving textures of less noise, particularly when utilizing reference feature injection.
We train the MLP with a constant learning rate of $0.001$ and LoRA weights for $\epsilon_\psi$ with a constant learning rate of $0.0001$ for a total of $4000$ iterations. The optimization is distributed over 4 GPUs with each sampling one view of the object. 

To meet the \textit{Identity Preservation} requirement, one could also use InstructPix2Pix~\cite{instructpix2pix} ($\mathrm{timbrooks/instruct{\text-}pix2pix}$) as in~\cite{instructnerf2nerf,instruct3dto3d} alternatively. 
We ablate such an option in~\cref{app:instructpix2pix} and find that InstructPix2Pix significantly limits versatility and generation quality, due to information loss incurred during their fine-tuning stage on a constrained instruction dataset. 
Additionally, we also consider a \textit{scene-agnostic texture editing} setup in~\cref{app:sceneagnosticediting} by converting the instructions into appearance descriptions.
The experiments reveal that our method is an effective alternative for general instruction-following 3D editing tasks, providing much more fine-grained and accurate control.

\paragraph{Light Estimation.} 
For post-processing for indoor scenes, we use $\mathrm{cv2.inpaint()}$ to fill holes and $\mathrm{cv2.morphologyEx()}$ to remove small isolated regions.
To locate the bright areas, we use $\tau_f=0.8$, $\tau_f=0.95$, $ \tau_o=0.9$, and $\tau_d=\mathrm{+inf}$. We set $\lambda_\mathrm{FOV}=1.65$ to include the potential full shadow. 

We use two \textit{global view}s only as input for vanilla SD\footnote{$\mathrm{stabilityai/stable{\text-}diffusion{\text-}2{\text-}base}$}: one always containing the full scene and one interpolated by cropping the full scene.
For light-dependent prompting, if the average intensity of the background areas
$\mathbb{E}(\{\mathcal{I}_{i,j} \ | \ i,j \notin$ \red{\faSunO} $\land \ i,j \notin$ \blue{\faLightbulbO}$\})<0.2$ and the average intensity of light areas $\mathbb{E}(\{\mathcal{I}_{i,j} \ | \ i,j \in$ \red{\faSunO} $\lor \ i,j \in$ \blue{\faLightbulbO}$\})<50.0$, we append ``\textit{in a dark environment}''.
Two viewpoints are forwarded to vanilla SD and supervised by VSD with the object prompt (with light-dependent prompting but without view-dependent prompting), a $\lambda=1.0$, a classifier-free guidance~\cite{classifierfree} $s_\mathrm{cf}=7.5$, and noise levels with $t\in[750, 990]$.
We train the light scales with a learning rate of $0.01$ and LoRA weights for $\epsilon_\psi$ with a learning rate of $0.0001$, all linearly annealed to $0.1\times$ through a total of $2000$ iterations. The optimization is distributed over 4 GPUs, with 3 each sampling one view of the object and the other sampling the sphere. We rescale the loss from each GPU with ratios $\{\frac{1}{6},\frac{1}{6},\frac{1}{6},\frac{1}{2}\}$ to balance the contributions from the object and the sphere. 

By default, the light estimation step is performed separately from the style adaptation step only to enhance the photorealism of the composition. One can perform light estimation first and reuse the estimated light during texture adaptation to disentangle potential lighting effects from being baked into textures. Our observation reveals that this yields less versatile texture especially when object parts are in shadow or strongly lit. 
However, for atypical lighting conditions such as dimly-lit scenes or environments with colored light, applying the estimated HDR maps during texture adaptation helps prevent the lighting effects from being baked into the texture map. Refer to~\cref{app:texturewithlight} for a detailed ablation study.

\section{Scene-Agnostic Texture Generation}
\label{app:sceneagnostic}

Without tailored design incorporating 2D scenes for mesh texturing, our pipeline effectively serves as an effective baseline on texturing textless meshes. We first detail its set-up in~\cref{app:sceneagnosticimplementation}, summarized as $\mathrm{(a)}$ in~\cref{tab:texturemetasetup}, and further compare it with the latest state-of-the-art methodologies~\cite{prolificdreamer,fantasia3d,texture,text2tex,texfusion} in~\cref{app:sceneagnosticcomparison}.

\subsection{Implementation Details}
\label{app:sceneagnosticimplementation}

\paragraph{Camera and Light Sampling.} 
We sample over $72$ views uniformly across azimuth $a\in[0^\circ,360^\circ]$ and elevation $e\in\{20^\circ,30^\circ,45^\circ\}$ with small perturbations. We observe that entirely random sampling in Mitsuba3~\cite{mitsuba3} leads to unknown memory leakage issues and rendering speed downgrade. We leave this for future fixes since it may require dedicated optimization of its compiler~\cite{drjit} from Mitsuba3's team.
Differently, we sample $\lambda_\mathrm{FOV} \in [0.6, 1.21]$ to include detailed close-up renderings of object parts, enhancing generation fidelity.
For light sampling, we use ambient lighting with the light scale being all $1.0$ by default and sample environment maps with one dominant light source by a probability $p$. Specifically, we model the light as an isotropic spherical Gaussian following~\cite{indoor}, parameterized by $c_x\in[0, 1.0]$, $c_y \in [0, 0.5]$, $c_r=0.08$, and $c_v\in[12.0, 15.0]$, denoting the x-coordinate of the Gaussian's center, y-coordinate of the Gaussian's center, radius, and its intensity. The background is simply parameterized by $b_v=0.8$ as its intensity. 
See coordinate conventions\footnote{\href{https://mitsuba.readthedocs.io/en/latest/src/generated/plugins\_emitters.html\#id2}{https://mitsuba.readthedocs.io/en/latest/src/generated/plugins\_emitters.html\#id2}} for the environment map used in Mitsuba3 for details. In practice, enabling environment map augmentation with a positive $p$ helps to alleviate the shortcut learning of reflectance as diffuse texture for highly reflective surfaces. 

\paragraph{Solid Color Composition.} We compose the object in the background with a default solid color $[0.5, 0.5, 0.5]$ and augment it with a probability $p=0.5$ with random solid colors. Alternatively, learning a background prediction network conditioned on camera poses is testified to be of little help. We discard the rendered shadow when composing it with the object rendering.

\paragraph{Textless Mesh.} 
We consider two main sources of textless meshes: (1) mesh generated from scratch~\cite{prolificdreamer,fantasia3d} with text prompts and SD guidance in Deep Marching Tetrahedra~\cite{dmtet} (DMTet); and (2) mesh downloaded from online assets~\cite{objaverse,polyheaven}.

\paragraph{Model, Prompt, and Guidance.} 
We use vanilla SD\footnote{$\mathrm{stabilityai/stable{\text-}diffusion{\text-}2{\text-}base}$} as the guidance diffusion model with a classifier-free guidance~\cite{classifierfree} of $7.5$. A $\lambda_t\in[0.9,1.0]$ is used to compare with other methodologies in~\cref{app:sceneagnosticcomparison}. However, we find that VSD tends to overfit 3D objects in a specific condition, such as lighting, leading to excessive noises and unnecessary details. 
Hence, it is sometimes essential to opt for an even smaller $\lambda_t$ (\ie, $0.6$) when \textit{scene-agnostic texture generation} setup is used as an initial stage for the texture adaptation step followed, leaving room to cultivate scene-conditional details, especially for simpler objects.
We use object prompts with view-dependent prompting as in~\cite{dreamfusion,prolificdreamer} but without light-dependent prompting, since we only sample properly-lit light conditions. Notably, we condition the LoRA-tuned model $\epsilon_\psi$ not only on light scales $c_v$ and $b_v$ but also on $c_x$, $c_y$, and $c_r$ as a 5-dimensional vector. For default ambient lighting, we set $c_x=0.25$ (\ie, $0^\circ$ azimuth), $c_y=0$ (\ie, $90^\circ$ elevation), $c_r=0$, $c_v=1.0$, and $b_v=1.0$. 

\paragraph{Stage-Wise Optimization.} By employing a coarse-to-fine 2-stage optimization, the pipeline strikes a balance between between performance and efficiency. For the first coarse stage, we use a rendering resolution of $256$ (and further resized to $512$) and samples per pixel of $64$. Noises $\epsilon$ with $t\in [30, 990]$ are sampled, ensuring a sufficient level of diversity. For the second fine stage, we use a rendering resolution of $512$ and samples per pixel of $128$. Noises $\epsilon$ with $t\in [500, 990]$ are sampled. A total of $4000$ iterations are conducted with training distributed across $4$ GPUs. Two stages are split with a ratio of $[0.2, 0.8]$. A constant learning rate of $0.001$ for the texture MLP and $0.0001$ for LoRA-tuning parameters are applied.

\subsection{Visual Comparison}
\label{app:sceneagnosticcomparison}

Comparison with two groups of methods: one based on \textit{texture wrapping}~\cite{texture,text2tex} and the other based on \textit{score distillation}~\cite{fantasia3d,prolificdreamer} are conducted. 

In~\cref{fig:sceneagnostic}, our method demonstrates superior photo-realistic texture generation, leveraging VSD.
Our results align closely with those of ProlificDreamer~\cite{prolificdreamer}, surpassing the comparatively cartoonish and saturated textures produced by Fantasia3D~\cite{fantasia3d}.
We note that it is imperative to use a small $\lambda$ for objects with simple textures (\eg, the \textit{umbrella}). Compared to ProlificDreamer which yields over-texturized texture, our method achieves reasonable texture for the \textit{umbrella}. 
Compared to TEXTure~\cite{texture} which generally lacks consistency over multiple views (\eg, the \textit{heart}), methods based on \textit{score distillation} provide a more cohesive and believable appearance.

\begin{figure*}
\centering
\includegraphics[width=\textwidth]{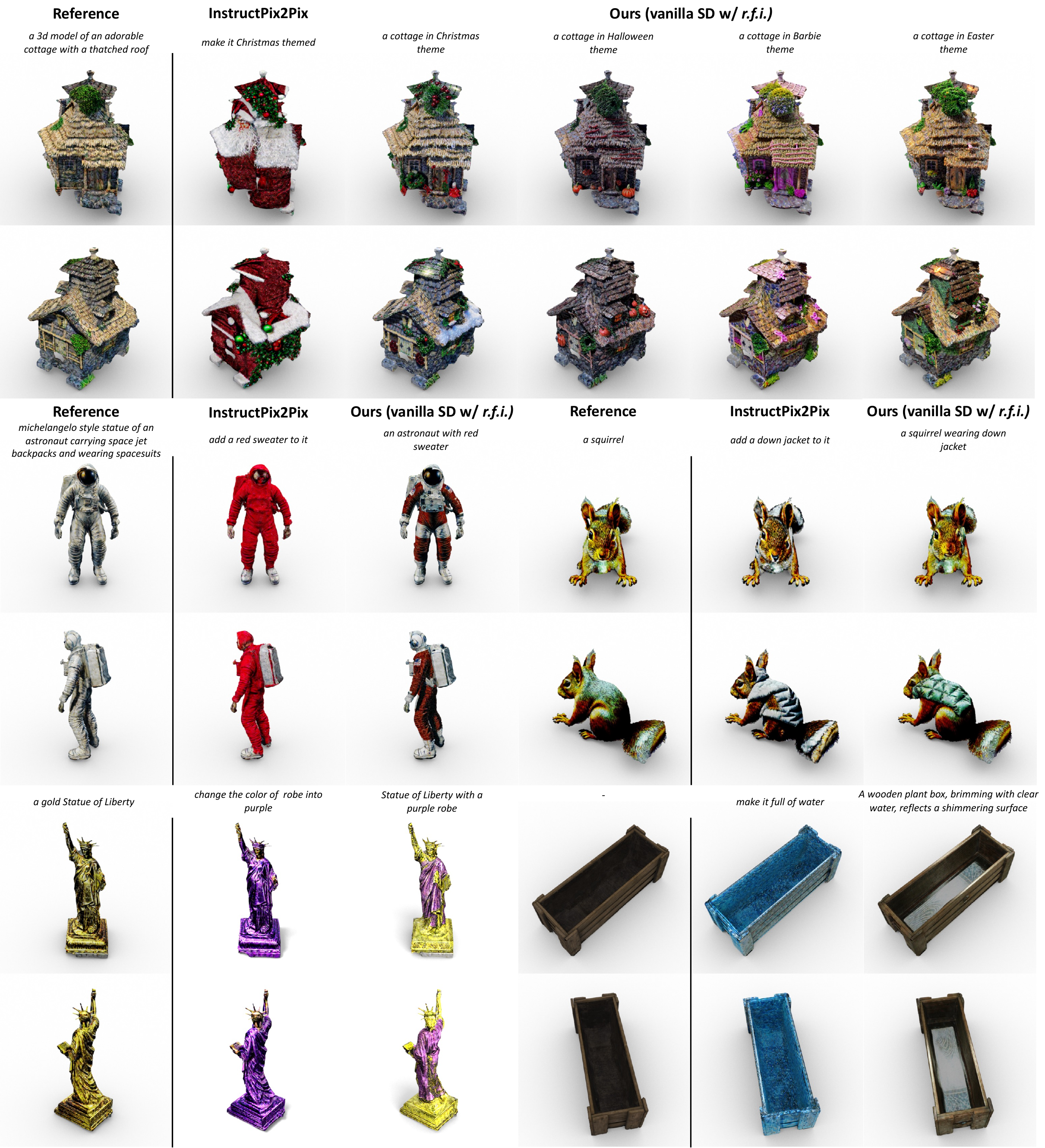}
\caption{\textbf{Scene-Agnostic Texture Editing.} 
We compare our framework with InstructPix2Pix~\cite{instructpix2pix} by replacing vanilla SD with reference feature injection (\textit{r.f.i.}). 
We use textured meshes generated from text prompts or existing downloaded textures as input.
}
\label{fig:sceneagnosticediting}
\end{figure*}

\section{Scene-Agnostic Texture Editing}
\label{app:sceneagnosticediting}

\subsection{Implementation Details}
\label{app:sceneagnosticeditingimplementation}
We follow the setup for \textbf{camera and light sampling} and \textbf{solid color composition} as in~\cref{app:sceneagnosticimplementation}. 
Since we are editing a textured mesh, we follow \textbf{textured mesh with neural texture} as in~\cref{app:textureimplementationdetails} to start from an initialized MLP weight, mitigating potential content drift.
We use the vanilla SD ($\mathrm{stabilityai/stable{\text-}diffusion{\text-}2{\text-}base}$) with reference feature injection as the guidance diffusion model with a classifier-free guidance~\cite{classifierfree} of $7.5$. We use a $\lambda_t=1.0$. We manually combine the object prompt (\eg, \textit{the leather sofa}) and the instruction (\eg, \textit{add sofa with dust}), yielding one describing the edited object's appearance (\eg, \textit{a leather sofa with dust}). Noises $\epsilon$ with $t\in [500, 990]$ are sampled with a rendering resolution of $512$ and samples per pixel of $128$. A total of $4000$ iterations are conducted with training distributed across $4$ GPUs. The setup is summarized as $\mathrm{(c)}$ in~\cref{tab:texturemetasetup}.

\subsection{Visual Comparison}
\label{app:sceneagnosticeditingcomparison}
We evaluate using InstructPix2Pix~\cite{instructpix2pix} as the guidance model and the instruction as text prompts, summarized as $\mathrm{(b)}$ in~\cref{tab:texturemetasetup}. Compared to InstructPix2Pix, vanilla SD with \textit{r.f.i.} achieves more fine-grained and accurate visual control. For example, while InstructPix2Pix successfully transforms the \textit{cottage} into a Christmas theme, it erases the object's original identity. Ours (vanilla SD with \textit{r.f.i.}) adds elements to the original \textit{cottage} (\eg, \textit{snow on the roof} and \textit{mistletoe at the top}), achieving a well-balanced integration of visual transfer and control.

\section{Prompting with GPT-4}
\label{app:gptprompt}

We detail the prompt used for prompting GPT-4 in~\cref{tab:gptprompt}. The generated prompts are denoted as \textit{GPT-prompted} in~\cref{tab:texturemetasetup} as opposed to the \textit{manually combined} ones (\eg, \textit{a leather sofa in a swamp}). 
We encourage GPT-4 to prioritize examining physical effects while leaving the consideration of lighting to the light estimation step. 
The follow-up instructions can be given from the users to provide more tailored details.
We encourage the user to first generate some 2D images using the text prompts to check if the output meets expectations.

\paragraph{Prompting for InstructPix2Pix~\cite{instructpix2pix}.} Similarly, to automatically generate scene-conditional text prompts for InstructPix2Pix to involve the \textit{Environmental Influence}, we use LLM to prompt for InstructPix2Pix, as studied in~\cref{app:instructpix2pix}.  
As detailed in~\cref{tab:gptpromptinstructpix2pix}, we prompt GPT-4 to give instructions instead of appearance descriptions for the final step after the analysis. The generated prompts are denoted as \textit{GPT-prompted} in~\cref{tab:texturemetasetup} as opposed to the \textit{manually instructed} ones (\eg, ``\textit{place the leather sofa in the swamp}''). 

\begin{table*}[!ht]
\centering
\scalebox{0.8}{
\begin{tabular}{p{0.15\linewidth}p{0.75\linewidth}}
\toprule
\multicolumn{2}{l}{\textbf{Prompting GPT-4 with optional visual input}} \\
\midrule
\ User & [the scene]: the given image (or \textit{a chaotic painting studio}) \\
& [the object]: \textit{a leather sofa} \\ \rule{0pt}{15pt}
& Now, imagine a picture where [the object] is placed into [the scene]. First, detailedly discuss the environmental impacts of [the scene] imposes on the appearance of [the object]. Focus on the physical impacts instead of lighting impacts. \\ \rule{0pt}{15pt}
& Then provide a succinct, purely descriptive, short, and mechanistic description of the appearance of [the object] inside [the scene] concatenated with ``,''. Aim the description to be approximately 25 to 35 words in length. Use simple language. Include all words provided in [the object]. \\ \rule{0pt}{15pt}
User (optional) & \raisebox{\dimexpr \topskip-\height}{\includegraphics[width=0.4\linewidth]{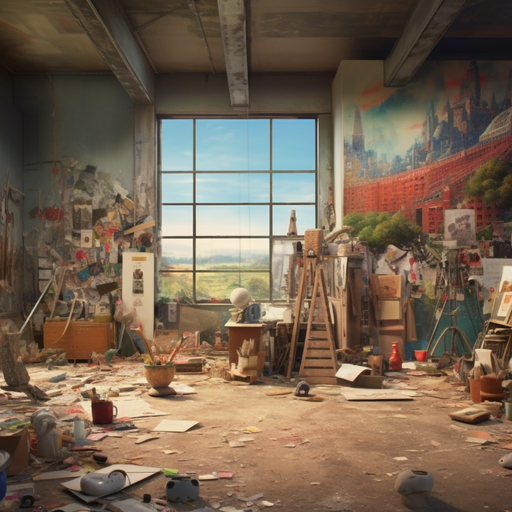}} \\ \rule{0pt}{15pt}
GPT-4 & In the given scene, which is a spacious and sunlit artist's studio cluttered with art supplies, canvases, and paint splatters, a leather sofa would become part of the creative tableau. The sofa would likely gather paint stains, dust, and perhaps even impressions from various tools and materials used in the studio. The natural lighting from the large windows could cause the leather to fade over time, especially if it's exposed to direct sunlight. \\ \rule{0pt}{15pt}
& The mechanistic description of the leather sofa in this scene would be: ``\textit{A leather sofa, stained with streaks and specks of paint, nestles amidst scattered canvases and brushes, its surface a testament to the studio's artistic endeavors.}'' \\ 
\midrule 
\ User & Let's suppose the leather sofa has been placed in the painting studio for decades. Consider the consequences and give me another description. \\ \rule{0pt}{15pt}
GPT-4 & \textit{A leather sofa, surface cracked and color faded, sits under a veil of dust and dried paint splatters, a silent witness to the studio's years of creative toil.} \\
\midrule
\ User & Let's suppose the leather sofa has just been placed. Consider the consequences and give me another description. \\ \rule{0pt}{15pt}
GPT-4 & \textit{A new leather sofa sits amidst the chaos of the painting studio, its pristine surface untouched yet, but poised to become a tapestry of paint splashes and creative wear over time.} \\
\midrule
\ User & Now, I want more paint on the sofa to make it look more messy. Consider the consequences and give me another description. 
 \\ \rule{0pt}{15pt}
GPT-4 & \textit{A leather sofa in the painting studio, now splattered with a kaleidoscope of paint, stands as a messy yet vibrant testament to the bustling creativity that surrounds it.} \\
\bottomrule
\end{tabular}}
\vspace{0.1cm}
\caption{Prompt used for prompting GPT-4 to generate a mechanistic description to reflect \textit{Environmental Influence} for Stable Diffusion~\cite{stablediffusion}. Note that the visual input is optional and the user can simply replace the given image with the scene prompt \textit{a chaotic painting studio}. 
The use can give follow-up context (\eg, how long the leather sofa is placed, more versus less paint, \etc) to achieve more precise visual control.
}
\label{tab:gptprompt}
\end{table*}

\begin{table*}[!ht]
\centering
\scalebox{0.8}{
\begin{tabular}{p{0.15\linewidth}p{0.75\linewidth}}
\toprule
\multicolumn{2}{l}{\textbf{Prompting GPT-4 with optional visual input}} \\
\midrule
\ User & [the scene]: the given image (or \textit{a chaotic painting studio}) \\
& [the object]: \textit{a leather sofa} \\ \rule{0pt}{15pt}
& Now, imagine a picture where [the object] is placed into [the scene]. First, detailedly discuss the environmental impacts of [the scene] imposes on the appearance of [the object]. Focus on the physical impacts instead of lighting impacts. \\ \rule{0pt}{15pt}
& Then provide a succinct instructions on how to change the appearance [the object] such that it looks realistic when placed in [the environment]. Start with each instruction with a verb and concatenate them with ``,''. Aim the description to be approximately 20 to 30 words in length. Use simple language. Include all words provided in [the object]. \\ \rule{0pt}{15pt}
User (optional) & \raisebox{\dimexpr \topskip-\height}{\includegraphics[width=0.4\linewidth]{pics/painting_studio_2.png}} \\ \rule{0pt}{15pt}
GPT-4 & Placing a leather sofa in the depicted artist's studio, which is cluttered and has a very lived-in appearance, would result in a number of physical changes over time due to environmental factors present in the scene. The leather on the sofa would likely become discolored and faded due to exposure to sunlight streaming in from the large window. Dust and paint splatters could accumulate on its surface, given the messy nature of the studio. If the studio is frequently used, the leather might show signs of wear and tear, such as cracks or creases, particularly on the seat and armrests where the user would have the most contact. If there are any liquids like turpentine or oils used in the studio, they could cause stains or deterioration of the leather. \\ \rule{0pt}{15pt}
& To make the leather sofa look realistic in this environment, the instructions would be: ``\textit{Age the leather, fade the color, add dust layers, simulate paint splatters, incorporate wear marks, show creases, and place stains strategically.}'' \\ 
\bottomrule
\end{tabular}}
\vspace{0.1cm}
\caption{Prompt used for prompting GPT-4 to generate a  succinct instruction to reflect \textit{Environmental Influence} for InstructPix2Pix~\cite{instructpix2pix}. Note that the visual input is optional and the user can simply replace the given image with the scene prompt \textit{a chaotic painting studio}. 
}
\label{tab:gptpromptinstructpix2pix}
\end{table*}

\section{Screenshots of User Study}
\label{app:userstudy}

\begin{figure}[!t]
    \centering
    \includegraphics[width=\linewidth]{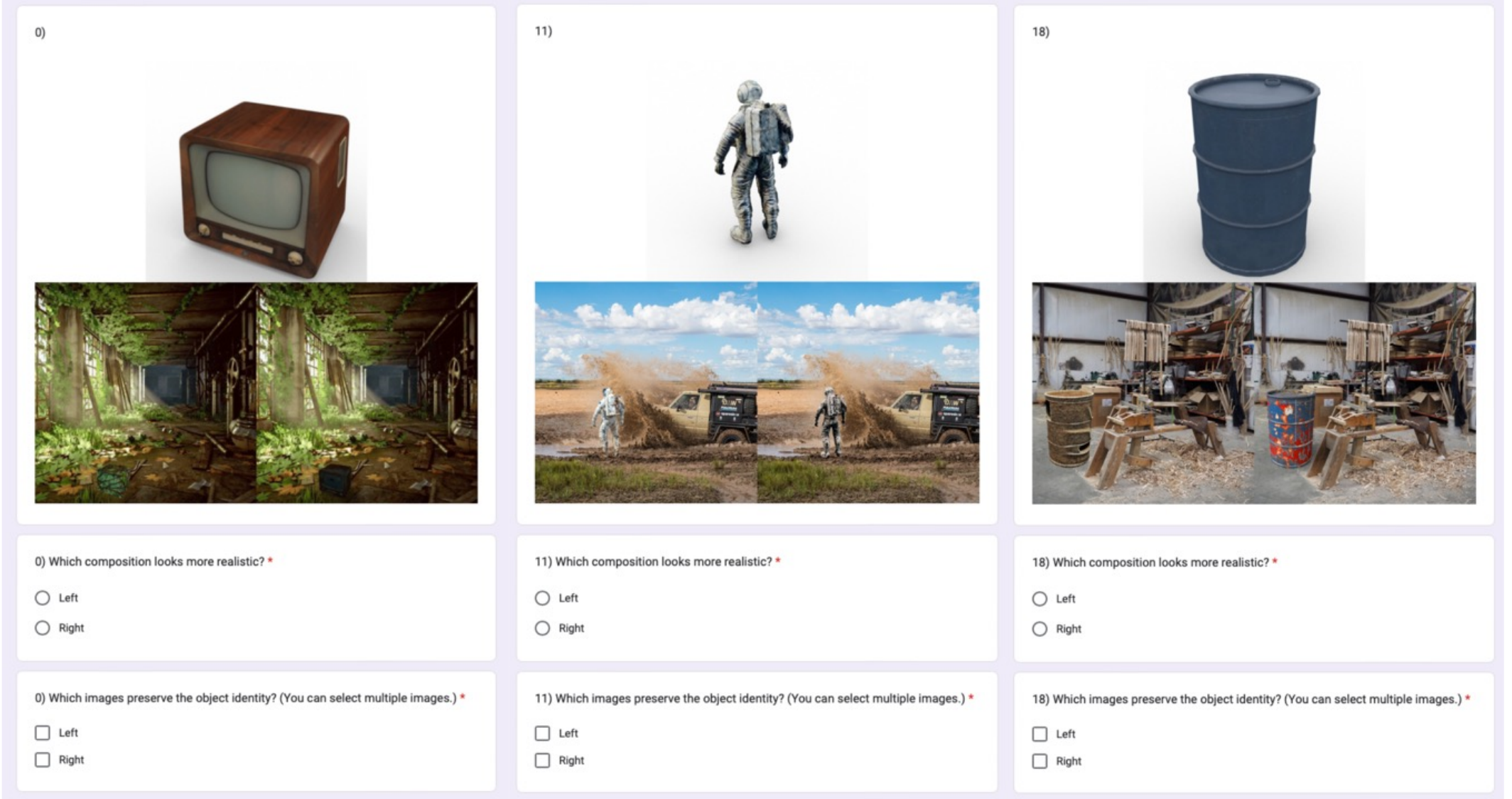}
    \vspace{-0.5cm}
    \caption{\textbf{Screenshots of User Study.} Users visually compared two images at a time, one from our method and another from a competing method, in random order.
    }
    \label{fig:userstudy}
\vspace{-0.4cm}
\end{figure}

In this section, we showcase screenshots captured during our user study. The study involved participants comparing images generated by different methods to evaluate their quality. Each participant was presented with pairs of images, one produced by our method and another by a competing method, in a random order. The aim was to gather subjective preferences to understand which method is perceived as superior (see Figure \ref{fig:userstudy}).

\section{Limitations and Potential Negative Impact}

\subsection{Limitations}

\paragraph{Complexity in Handling Diverse Environments.} While the paper showcases success across a variety of indoor and outdoor scenes, the complexity and unpredictability of real-world environments may present challenges. The method might struggle with scenes that have highly complex lighting conditions or where the environment heavily influences the object's appearance.

\paragraph{Dependence on Accurate LHR Estimation.} 
The framework's effectiveness in light estimation hinges on the precise estimation of the LHR map. Any inaccuracies in these estimations could lead to sub-optimal HDR map estimation and thus undesirable object-scene composition.

\paragraph{Scalability and Efficiency.} The optimization process relies on employing differentiable ray tracing and conditioning on diffusion models, which is computationally intensive. This could limit the method's scalability or applicability in real-time or resource-constrained scenarios.

\subsection{Potential Negative Impact}

\paragraph{Misuse in Creating Deceptive Media.} The ability to seamlessly integrate and adapt 3D objects into 2D scenes can be misused to create realistic yet deceptive images. This technology could contribute to the proliferation of deepfakes or other forms of misleading content, impacting areas like news media, legal evidence, and personal security.

\paragraph{Intellectual Property Infringement.} The framework enables the realistic integration of 3D models into various 2D scenes, which could lead to unauthorized use of copyrighted or trademarked objects within new contexts. This poses potential concerns regarding intellectual property rights and could facilitate copyright infringement.


\paragraph{Acknowledgement.} 
This work is supported by the UKRI grant: Turing AI Fellowship EP/W002981/1 and EPSRC/MURI grant: EP/N019474/1.
J. Zhou is also supported by the Horizon Robotics.
We thank Luke Melas-Kyriazi and Fabio Pizzati for their helpful discussions.

%
%
\bibliographystyle{splncs04}
\bibliography{egbib}

\end{document}